\documentclass[10pt,technote,compsoc]{IEEEtran}
\usepackage{amsmath,amsfonts}
\usepackage{algorithm}
\usepackage{array}

\usepackage[caption=false,font=small,labelfont=normalfont,textfont=normalfont]{subfig}

\usepackage{caption}
\usepackage{textcomp}
\usepackage{stfloats}
\usepackage{url}
\usepackage{verbatim}
\usepackage{graphicx}
\usepackage{cite}
\usepackage{colortbl}
\usepackage{pifont}
\usepackage{textcomp}
\usepackage{stfloats}
\usepackage{threeparttable}
\usepackage{url}
\usepackage{verbatim}
\usepackage{graphicx}
\usepackage{cite}
\usepackage{diagbox}
\usepackage{bm}
\usepackage{bbm}
\usepackage{soul}
\usepackage{amsfonts}
\usepackage[breaklinks,colorlinks]{hyperref}
\usepackage{multirow}
\usepackage{algpseudocode}
\usepackage{makecell}
\usepackage{booktabs}
\usepackage{mdframed}
\usepackage[dvipsnames]{xcolor}
\usepackage{booktabs}
\usepackage{bbding}
\usepackage[top=0.48in,bottom=0.48in,left=0.39in,right=0.39in]{geometry}
\usepackage{array}
\usepackage{ulem}
\usepackage{ragged2e}
\definecolor{mycolor}{RGB}{169,189,211}

\newcommand{\etal}{\textit{et al}. }
\newcommand{\etc}{\textit{etc}. }
\newcommand{\ie}{\textit{i}.\textit{e}., }
\newcommand{\eg}{\textit{e}.\textit{g}., }
\newcommand{\wrt}{\textit{w}.\textit{r}.\textit{t}. }

\begin{document}
	
	\title{Enhancing Transferability of Targeted Adversarial Examples: A Self-Universal Perspective}
	
	\author{Bowen Peng, Li Liu, Tianpeng Liu, Zhen Liu, Yongxiang Liu  
		\IEEEcompsocitemizethanks{
			\IEEEcompsocthanksitem The authors are with the College of Electronic Science and Technology, National University of Defense Technology (NUDT), Changsha 410073, China. Email: pbow16@nudt.edu.cn, liuli\_nudt@nudt.edu.cn, everliutianpeng@sina.cn, zhen\_liu@nudt.edu.cn, lyx\_bible@sina.com.
		}
		\thanks{Corresponding Authors: Li Liu and Yongxiang Liu.}
		\thanks{This work was supported partially by the National Key Research and Development Program of China under Grant 2021YFB3100800, the National Natural Science Foundation of China under Grant 62376283, 61921001, 62022091, and 62201588.}
	}

        \markboth{Preprint by IEEE \LaTeX~Template}{}

	\IEEEtitleabstractindextext{
		\vspace{-3mm}
		\begin{abstract}
			Transfer-based targeted adversarial attacks against black-box deep neural networks (DNNs) have been proven to be significantly more challenging than untargeted ones. The impressive transferability of current SOTA, the generative methods, comes at the cost of requiring massive amounts of additional data and time-consuming training for each targeted label. This results in limited efficiency and flexibility, significantly hindering their deployment in practical applications. In this paper, we offer a self-universal perspective that unveils the great yet underexplored potential of input transformations in striking an effective balance between transferability, efficiency, and flexibility. Specifically, transformations universalize gradient-based attacks with \textit{intrinsic but overlooked semantics} inherent within \textit{individual images}, exhibiting similar scalability and comparable results to time-consuming learning over massive diverse data. We also contribute a surprising empirical insight that one of the most fundamental transformations, simple image scaling, is highly effective, scalable, sufficient, and necessary in enhancing targeted transferability. We further augment simple scaling with orthogonal transformations and blockwise applicability, resulting in the Simple, faSt, Selfuniversal yet Strong Scale Transformation (S$^4$ST) for self-universal transferable targeted attacks. On the ImageNet-Compatible benchmark dataset, our method achieves a 19.8\% improvement in the average targeted transfer success rate against various challenging victim models over existing SOTA transformation methods while only consuming 36\% time for attacking. It also outperforms resource-intensive attacks by a large margin in various challenging settings. Codes are available at \url{https://github.com/scenarri/S4ST}.
		\end{abstract}
		\vspace{-5mm}
		\begin{IEEEkeywords}
			\justifying 
			Adversarial machine learning, gradient-based attack, targeted attack, transferability 
	\end{IEEEkeywords}
}
	\vspace{-5mm}
	
	\maketitle
	\IEEEdisplaynontitleabstractindextext
	\IEEEpeerreviewmaketitle
	
	\section{Introduction}
	Since Adversarial Examples (AEs) and their transferability were identified \cite{szegedy2013intriguing,harnessing2015goodfellow}, the pursuit of more transferable AEs has fostered ongoing advances in understanding and defending modern artificial intelligence \cite{madry2018towards,tramer2018ensemble,10214340,10478545}. Compared with only falsifying black-box models' predictions \cite{SIA,BSR,decowa,momentum2018dong,evading2019dong}, directing them towards a targeted label offers particular benefits for privacy and intellectual property protection \cite{chen2023selfensemble} and the deployment of neural networks in real-world and safety-critical applications \cite{ODI,9858024}, \etc Despite a decade-long research endeavor \cite{liu2016delving,PoTrip,logit,SelfU,AOSBLL,marginangleT}, achieving Transferable Targeted Attacks (TTAs) remains a formidable challenge, with difficulty arises from the complexities involved in crafting target label-aligned semantics (or features) \cite{Ilyas19,domainfeatureuap,naseer2021generating} that are robust against model discrepancies \cite{ben2006analysis} while adhering to constraints for visual imperceptibility.
	
	The existing TTA methods are primarily based on two distinct principles \cite{logit}. \textit{Simple TTAs} focus on conducting gradient descent over a target sample-label pair \cite{logit,marginangleT,PoTrip}. While being efficient and flexible, they suffer from overfitting to the surrogate model, which can significantly hinder successful transfers, frequently leading to near-zero success rates without the application of image transformations \cite{inputdiversity2019xie,RDI,ODI}. \textit{Resource-intensive TTAs} harness the victim models' training data to facilitate targeted transferability \cite{naseer2021generating,zhao2023minimizing,domainfeatureuap,weng2023exploring}. Their effectiveness stems from harnessing the underlying data distribution and exhibiting universality towards unseen samples. However, the practicality of these attacks is hampered by their reliance on substantial data volumes and the necessity for extensive retraining when targeting new labels, which detracts from their efficiency and flexibility.
	
	\begin{figure}[tbp]
		\centering
		\vspace{-2mm}
		\subfloat[\footnotesize Results by SOTA method, M3D \cite{zhao2023minimizing} (resource-intensive TTA). When the generator is more universal to to more additional samples, it exhibits stronger transferability while inevitably consuming formidable time for training.
		]{\includegraphics[width=1\linewidth]{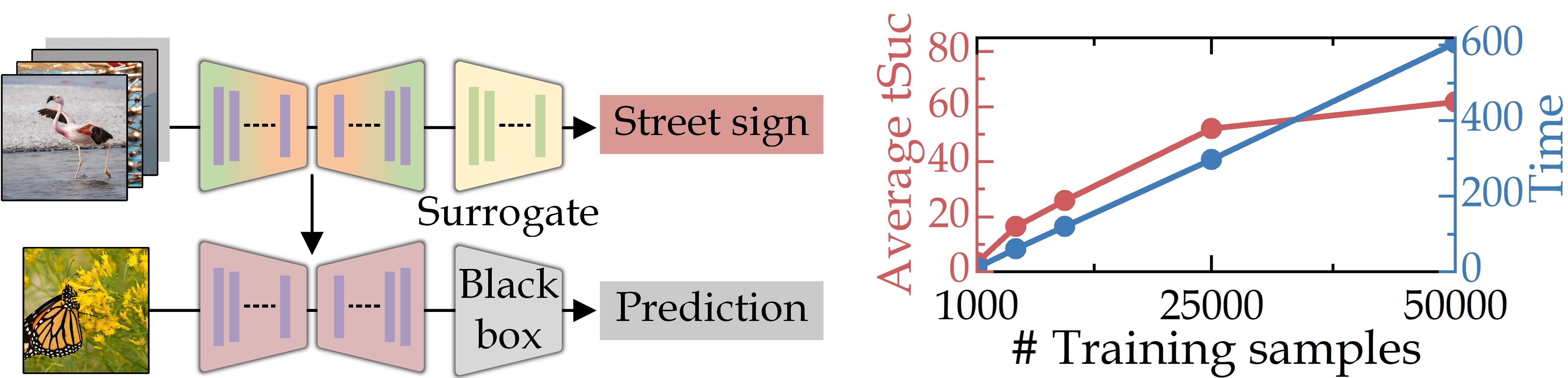}\vspace{-9mm}} \vspace{-3mm}\\
		\subfloat[\footnotesize Results by TI-MI-FGSM (TMI) \cite{evading2019dong,momentum2018dong}  with simple scaling transformation (simple TTA). When the perturbation is more universal to more transformed self-copies, it exhibits stronger transferability while maintaining consistent efficiency.
		]{\includegraphics[width=1\linewidth]{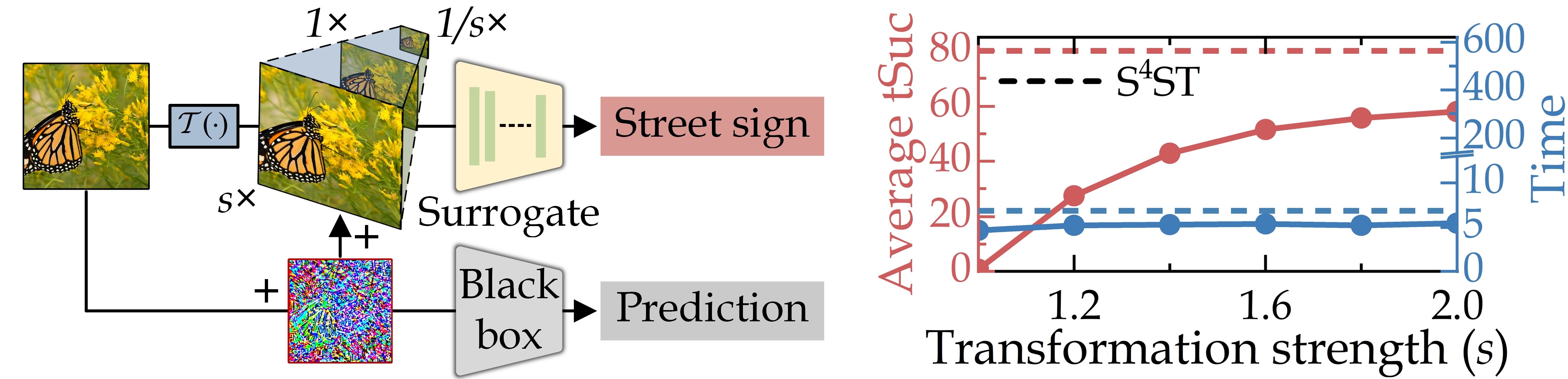}\vspace{-9mm}}
		\vspace{-1mm}
		\caption{Comparison of targeted transfer success rate (tSuc) and time consumption for training and attack (minutes), averaged over 14 black-box models and 1,000 images. The surrogate model is ResNet-50, and the target label is \texttt{street sign}. Our \textit{self-universal} hypothesis collaborates (a) the universality to additional samples and (b) the self-universality to transformed copies of a single sample, motivating us to find a potent transformation for flexible, efficient, and powerful self-universal TTA. We are particularly interested in the simple scaling transformation, which we empirically demonstrate as scalable, sufficient, and necessary for enhancing targeted transferability. We further propose the S$^4$ST transformation, which surpasses both SOTA transformation methods and resource-intensive methods; please see Figs. \ref{transformations} and \ref{t-s-plot}.}
		\label{illu}
		\vspace{-7mm}
	\end{figure}
	
	In this paper, our objective is to address the following question: \textit{``Can we attain superior performance surpassing resource-intensive methods solely by utilizing the target sample itself, thereby achieving the best of both worlds?''} Our focus is directed toward the role of input transformations which, while inherently critical within simple TTAs, have hitherto been underexplored. Our central insight is the \textit{self-universal} hypothesis, with motivation provided by Fig. \ref{illu} and details in Section \ref{suh}. In brief, transformations universalize data-specific optimization by activating and manipulating the \textit{intrinsic semantics} inherent within \textit{individual images} but \textit{overlooked} by trained models, thus enabling and enhancing the transferability of simple TTAs, which shares a similar spirit to harnessing large volumes of multi-class data. It unveils a direct way to achieve our objective: by scaling and expanding transformation methods, akin to scaling the training samples of resource-intensive methods, which we refer to as self-universal TTA. This simple assumption actually brings a new perspective to understanding how transformations benefit transferability, as it directly contradicts previous guiding principles for transformation design, especially for untargeted attacks, where transformations need to be constrained to a small strength to be loss-preserving \cite{nestrov2019lin,evading2019dong}. 
	
	To facilitate self-universal TTA, we commence with one of the most fundamental transformations, the simple image scaling. This choice is informed by both human visual perception, where different objects are prominent at various scales, and the analogous traits observed in deep neural networks \cite{Yun_2021_CVPR}. We show that simple scaling can surpass complex methodologies engineered based upon it \cite{inputdiversity2019xie,RDI}. Further, tactful experiments were conducted to uncover the scalability, sufficiency, and necessity of simple scaling to enhance targeted transferability, suggesting semantics at varying scales serve as a good proxy for leveraging the inherent but overlooked semantics.
	
\begin{figure*}[tbp]
		\centering
		\subfloat[\footnotesize  Images transformed by DI \cite{inputdiversity2019xie}, RDI \cite{RDI}, ODI \cite{ODI}, SI \cite{nestrov2019lin}, Admix \cite{admix}, SSA \cite{SSA}, DeCoWa \cite{decowa}, SIA \cite{SIA}, BSR \cite{BSR}, and our S$^4$ST.\label{existing}]{\includegraphics[width=0.478\linewidth]{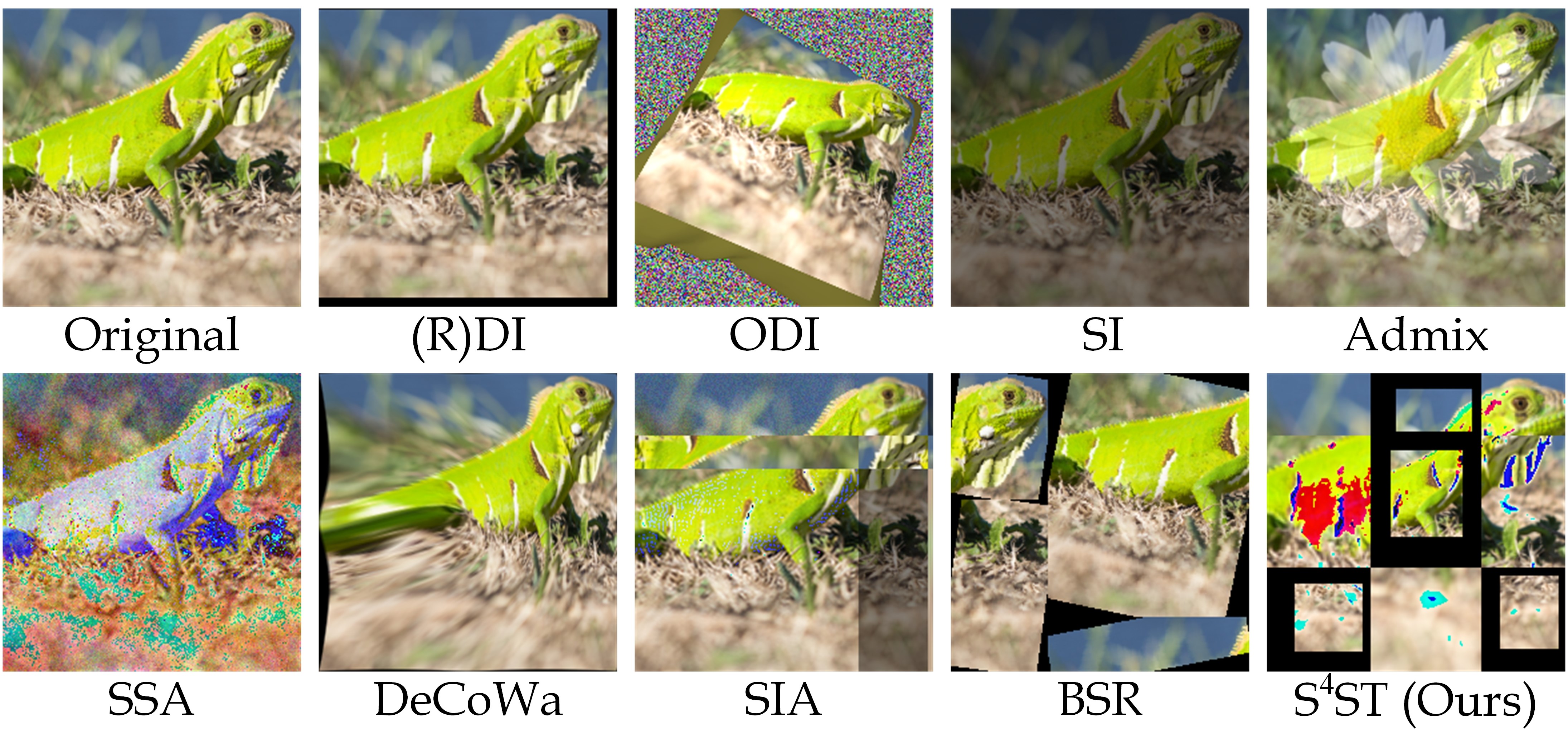}\vspace{-7mm}} \hspace{2mm}
		\subfloat[\footnotesize The proposed S$^4$ST consists of the \textit{Base} operation, which founded on scaling, and two orthogonal enhancements—\textit{Aug} and \textit{Block}.\label{s4st}]{\includegraphics[width=0.478\linewidth]{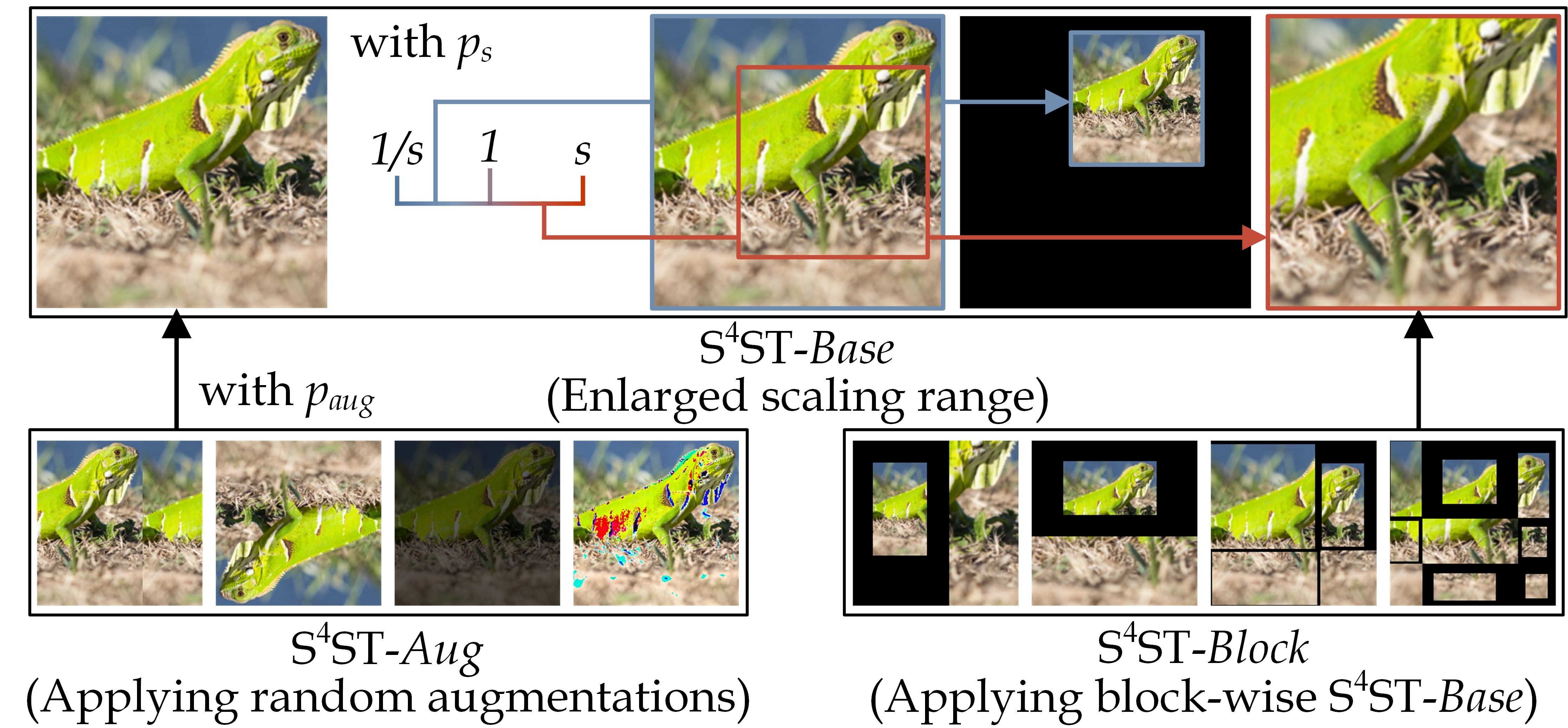}\vspace{-7mm}}
		\vspace{-1mm}
		\caption{Illustration of the original image and its transformed copies by existing methods and our S$^4$ST.}
		\label{transformations}
		\vspace{-7mm}
	\end{figure*}

\begin{figure}[tbp]
		\includegraphics[width=1\linewidth]{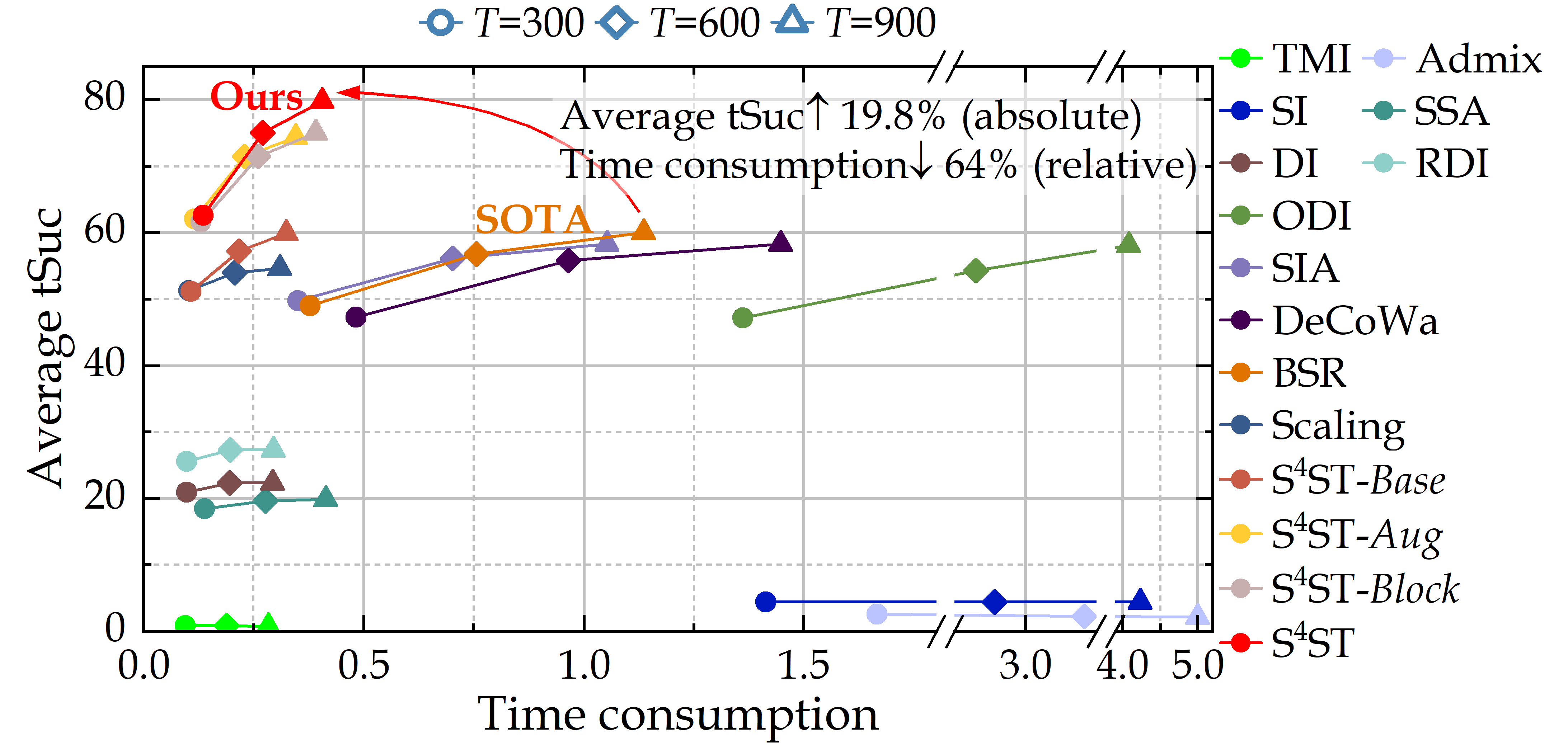}
		\vspace{-6mm}
		\caption{Comparison of the average tSuc and time consumption (seconds) to craft an AE on the ImageNet-Compatible dataset. The results are obtained using TMI as the baseline attack with a perturbation budget of $\ell_{\infty}=16/255$ at different iterations $T$. Our S$^4$ST exhibits superior effectiveness and efficiency over existing transformations. Compared with SOTA transformation, BSR \cite{BSR}, S$^4$ST yields an absolute improvement of 19.8\% on average tSuc to 79.8\%, and a relative reduction of 64\% on time consumption to 0.41s.}
		\label{t-s-plot}
		\vspace{-6mm}
	\end{figure}

	Grounded in above findings, we further devise a Simple, faSt, Self-universal yet Strong Scale Transformation (S$^4$ST) for TTA. As shown in Fig. \ref{s4st}, S$^4$ST integrates two efficient strategies, distinct from but inspired by recent advancements \cite{SIA,BSR}, to increase diversity further: \textit{1)} collaboration with random orthogonal transformations; and \textit{2)} application of block-wise scaling. The proposed S$^4$ST transformation is evaluated comprehensively and proven to be both efficient and effective in facilitating targeted transferability with results outlined in Fig. \ref{t-s-plot}. 
	
	In summary, our main contributions are as follows:
	\vspace{-1.5mm}
	\begin{itemize}
		\item We contribute a novel and concise self-universal perspective, shedding new light on elucidating the pivotal role of input transformations in TTA. In contrast to the previous consensus, it reveals the feasibility of enhancing targeted transferability by \textit{strong} transformations.
		\item To the best of our knowledge, we are the \textit{first} to experimentally investigate the impact of a single transformation method on targeted transferability. Our empirical demonstration highlights the significant potential of simple scaling in enhancing TTA. Furthermore, we propose the S$^4$ST to integrate complementary gains from existing methods to achieve superior performance.
		\item We conduct comprehensive experiments against diverse and challenging victim models under various settings on the ImageNet-Compatible benchmark dataset. Results demonstrated that our method outperforms SOTA methods by a significant margin in terms of both effectiveness and efficiency.
	\end{itemize}

	\vspace{-5mm}
	\section{Related Works}\label{pre}
	
	We follow Zhao \etal \cite{logit} to categorize current TTA methods into simple and resource-intensive ones.
	
	\vspace{-3mm}
	\subsection{Simple TTA}	
	\vspace{-1mm}
	Simple TTA performs gradient-based iterative optimization over a target sample-label pair based on a surrogate model. The existing method for untargeted transferable attacks can be directly utilized for the targeted case, such as I-FGSM \cite{kurakin2016adversarial} and MI-FGSM \cite{momentum2018dong}. However, the well-established negative cross-entropy loss, $\mathcal{L}_{\text{NCE}}$, encounters challenges in TTA due to the issue of gradient vanishing. To this end, Li \etal proposed a Poincaré distance-based triplet loss, $\mathcal{L}_{\text{Po+Trip}}$, for effective optimization \cite{PoTrip}. The Logit loss by Zhao \textit{et al.} \cite{logit}, $\mathcal{L}_{\text{Logit}}$, directly maximizes the target logit, serves as a straightforward way to bypass the gradient vanishing problem. There are several followup studies tried to soften the $\mathcal{L}_{\text{NCE}}$ to stabilize the optimization, \eg by margin-based logit calibration \cite{marginangleT}, $\mathcal{L}_{\text{NCE}}^{\text{MLC}}$, or adversarial optimization scheme \cite{AOSBLL}.
	
	Another line of research focuses on input transformation. Analogous to data augmentation for model training, these methods transform the samples at each attack iteration. The Diverse Inputs (DI) method \cite{inputdiversity2019xie} has been identified as pivotal in simple TTA \cite{logit}. The optimization process required substantially more iterations ($\sim$300) to converge compared to the untargeted setting ($\sim$10). Further, DI's mild variation Resized DI (RDI) \cite{RDI} was found to be more effective in \cite{CFM}. Subsequently, the (R)DI method became a standard baseline for later research.  Byun \etal introduced the Object-based DI (ODI) \cite{ODI} method, which is particularly designed for TTA, projecting the image onto 3D objects and utilizing renderings from diverse viewpoints for inference. This work also includes advanced transformation methods designed for untargeted transferable attacks. We list involved methods in Fig. \ref{existing}.

	\vspace{-5mm}
	\subsection{Resource-Intensive TTA}
	\vspace{-1mm}
	
	Resource-intensive TTA is not confined to any specific paradigms. However, they all attempt to enhance targeted transferability by harnessing data distribution learned from an extensive amount of additional data. Targeted universal attacks \cite{domainfeatureuap} train one single perturbation over massive data, thus being agnostic to data and can attack multiple unseen samples, exhibiting adequate cross-model transferability. Inkawhich \etal opt to train auxiliary models at intermediate layers to learn and further attack the feature distribution \wrt the target class \cite{inkawhich2020perturbing}.	
	The current SOTA of TTA is achieved by generative methods. The first representative generative framework for TTA is Transferable Targeted Perturbation (TTP) \cite{naseer2021generating}, where a generator is trained over massive data from diverse source classes, also equipped with dense transformation, to match the target class in both global distribution and local neighborhood structure. Wang \etal \cite{wang2023towards} combined TTP and \cite{inkawhich2020perturbing}, as well as proposed a perturbation dropping strategy to boost performance further. Zhao \etal proposed the Minimizing Maximum Model Discrepancy (M3D) attack \cite{zhao2023minimizing}, which concurrently differentiates and attacks two surrogate models (the original and its copy), achieving current SOTA performance. Though effective and efficient at test, the applicability of these methods is largely limited by the need for massive training data from the same domain of victims and time-consuming retraining per label.
	
	\vspace{-5mm}
	\section{The Proposed Methodology}\label{sec_method}
	\vspace{-1mm}
	\subsection{Preliminary}
	\vspace{-1mm}
	Consider a trained image classifier $F$ that predicts among $K$ categories, $\bm{y}=\{y_i|y_i\in\mathbb{Z},0\leq i<K\}$, for RGB-channel input, $\bm{x}\in \mathbb{R}^{3 \times H \times W}$, where $H$ and $W$ are respectively the image height and width. Without loss of generality, we denote $F=C \circ f$, where $f(\bm{x})=\bm{p}\in[0,1]^{K}$ is the confidence score for each category, and $C(\bm{p}, k) = y_{\operatorname{argsort}(\bm{p}, k)}$ indexes the corresponding label for the $k$-th highest prediction (by default, $k=1$). TTA aims to force an arbitrary target prediction, $y_t$, at arbitrary inputs for a black-box classifier, $G=C\circ g$:
	\vspace{-2mm}
	\begin{equation}
		\label{obj}
		\centering
		G(\bm{x}^{adv}) = y_t, \quad \textit{s.t.} \quad  \|\bm{x}^{adv}-\bm{x}\|_{p} \leq \epsilon,
		\vspace{-2mm}
	\end{equation}
	where $\epsilon$ denotes the perturbation budget. In the transfer attack setting, $\bm{x}^{adv}$ is crafted based on $f$ (surrogate model):
	\vspace{-2mm}
	\begin{equation}
		\centering \bm{x}^{adv}=\underset{\bm{x}^{\prime}:\left\|\bm{x}^{\prime}-\bm{x}\right\|_p \leq \epsilon}{\arg \max } \mathcal{L}\left(f(\bm{x}^{\prime}), y_t\right),
		\vspace{-2mm}
	\end{equation}
	where $\mathcal{L}$ is a loss function. The above problem is generally solved by input transformation-empowered TMI with $\ell_{\infty}$ constraint for $T$ steps: 
	\vspace{-2mm}
	\begin{equation}
		\centering
		\label{TMI_eq}
		\begin{aligned}
			&\bm{x}^{adv}_0 = \bm{x}, \quad \bm{g}_0 = \bm{0},\\
			\bm{g}_{i+1}=\mu \cdot \bm{g}_i&+\frac{\bm{W}*\nabla_{\bm{x}} \mathcal{L}(f(\mathcal{T}(\bm{x}^{adv}_i), y_t))}{\left\|\bm{W}*\nabla_{\bm{x}} \mathcal{L}(f(\mathcal{T}(\bm{x}^{adv}_i), y_t))\right\|_1},\\
			\bm{x}^{adv}_{i+1} &= \Pi_{\epsilon}(\bm{x}^{adv}_{i}+\alpha \cdot \operatorname{sign} (\bm{g}_{i+1})),
		\end{aligned}
	\end{equation}
	where $\alpha$ is step size, $\Pi_{\epsilon}$ is the projection operation, $\mathcal{T}$ denotes a input transformation method, $\mu$ is the decay factor for momentum accumulation \cite{momentum2018dong}, and $\bm{W}$ is a predefined kernel for gradient convolution \cite{evading2019dong}.
	
	\vspace{-4mm}
	\subsection{Self-Universal Hypothesis}\label{suh}
\begin{figure}[tbp]
		\includegraphics[width=1\linewidth]{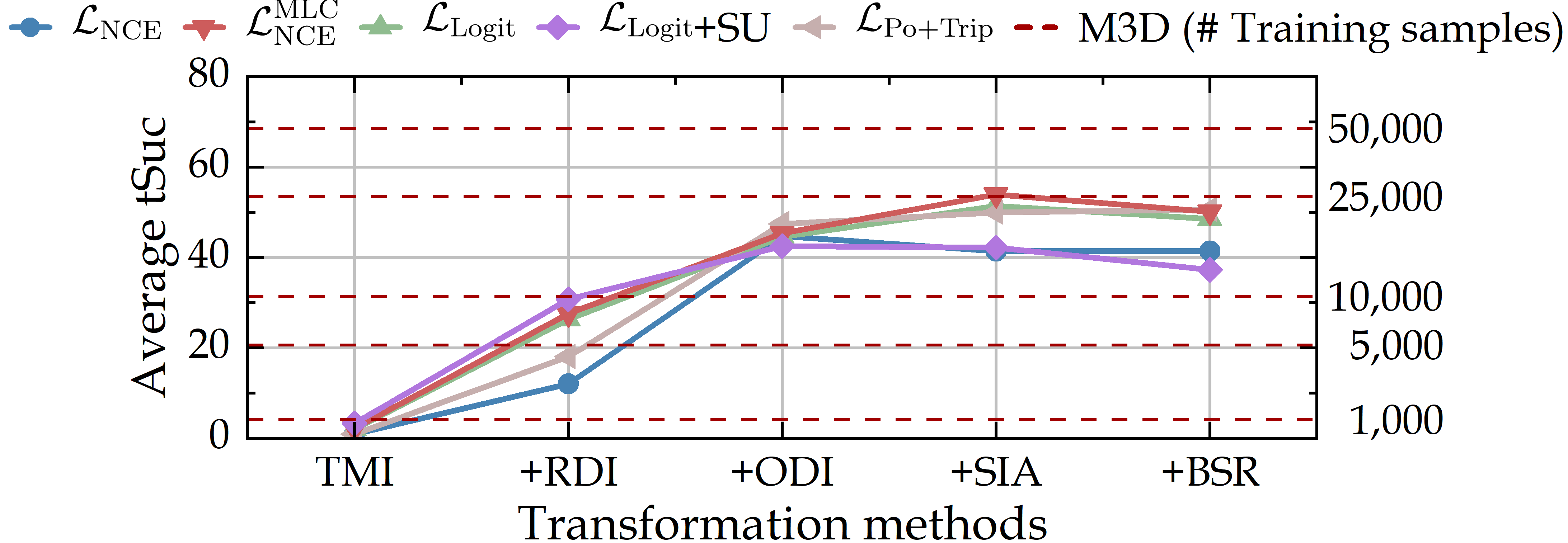}
		\vspace{-5mm}
		\caption{Average tSuc evaluated on the ImageNet-Compatible dataset under 10-Targets (all-source) setting \cite{naseer2022stylized}. Input transformations are identified as playing an indispensable role in enabling effective simple TTA. Moreover, the transformation strength for simple TTA and the training sample volumes for resource-intensive TTA exhibit similar impacts on attack performance.}
		\label{loss_trans}
		\vspace{-6mm}
	\end{figure}

	\vspace{-1mm}
	\textbf{Observation from Existing Methods}. Current research primarily focuses on the design of loss functions and transformation methods. However, these two categories of progress lack complementary evaluation. We commence with assessing existing methods, \ie implementing Eq. \eqref{TMI_eq} with different $\mathcal{L}$ and $\mathcal{T}$. Following the previous setting, we set $\alpha=2/255$, $\epsilon=16/255$, and $T=300$ for simple TTAs, and we also reproduce M3D with different numbers of training samples. The results are shown in Fig. \ref{loss_trans}. Notable improvements in tSuc are only achieved when applying input transformations, with minor variations observed across different loss functions. Importantly, it exhibits a striking resemblance to the impact of training sample volumes on resource-intensive methods—both lack of transformation and sufficient training samples result in near zero tSuc, more potent transformation, and more training samples will yield significant improvements.
	
	\textbf{The Self-Universal Hypothesis}. Observations highlighted above reveal the pivotal role of input transformations in enabling effective simple TTA, akin to utilizing multi-class data by resource-intensive methods. This motivates us to propose the self-universal hypothesis, positing that exploiting the transformation-induced multi-class information within individual images is similar to the distributional knowledge of multi-class data in enhancing TTA. Specifically, if a perturbation is \textit{self-universal} to transfer \textit{all semantics} inherent within an image toward the target class, it should be more transferable to attack other black-box models. Here, \textit{all semantics} particularly emphasize the \textit{subsistent} semantics present within individual images but \textit{overlooked} during general inference, as will be shown in Fig. \ref{confidence_score_illu}. Our self-universal hypothesis unveils a straightforward avenue to enhance the targeted transferability: by employing sufficiently strong transformations to activate and attack \textit{more} semantics that have been usually neglected. Through such transformations, our goal is to achieve highly effective simple TTA that outperforms the resource-intensive methods while maintaining flexibility in handling arbitrary sample-label pairs without needing additional data or model training, thus achieving the best of both worlds.
	
	\textbf{Discussion}. \textit{1)} Our analysis, in fact, contradicts the initial design criteria for transformation methods, which necessitate a loss-preserving transformation rather than promoting diversity of features \cite{evading2019dong,inputdiversity2019xie}. However, recent works increasingly demonstrate that a transformation with stronger diversity leads to better transferability, whether in untargeted or targeted attacks, albeit requiring more gradient computations to stabilize the optimization process \cite{SIA,BSR,decowa}. We demonstrate in Fig. \ref{t-s-plot} that compared to previous settings ($T=300$), additional optimization iterations yield significant improvements to strong transformations. \textit{2)} A similar nomenclature, the self-university, can be found in \cite{SelfU} by Wei \etal Essentially, our motivation aligns—we both aim to maximize the utilization of knowledge from a single image to enhance targeted transferability. Nonetheless, our focus diverges significantly as we concentrate on introducing a more effective transformation to achieve this goal while they add an extra inference branch to pursue consistent adversarial effect between random crops and the whole image. We implement the method of Wei \etal in our setting, and the results are provided in Fig. \ref{loss_trans} by $\mathcal{L}_{\text{Logit}}+\text{SU}$.

	\vspace{-4mm}
	\subsection{Simple Scaling to Boost Targeted Transferability}
	\vspace{-1mm}
	\textbf{Motivation}. There are two mainstreams of input transformation methods: \textit{1)} dynamically searching optimal combinations amongst a vast pool of transformations during the attack iteration \cite{yuan2022adaptive,Zhu_2024_CVPR}; and \textit{2)} applying a comprehensively defined transformation in each iteration \cite{inputdiversity2019xie,evading2019dong,decowa}. In this work, we opt to identify a foundational transformation that is sufficient and necessary for self-universal TTA, upon which we plan to carry out complementary enhancements.
	
	\begin{figure}[tbp]
		\includegraphics[width=1\linewidth]{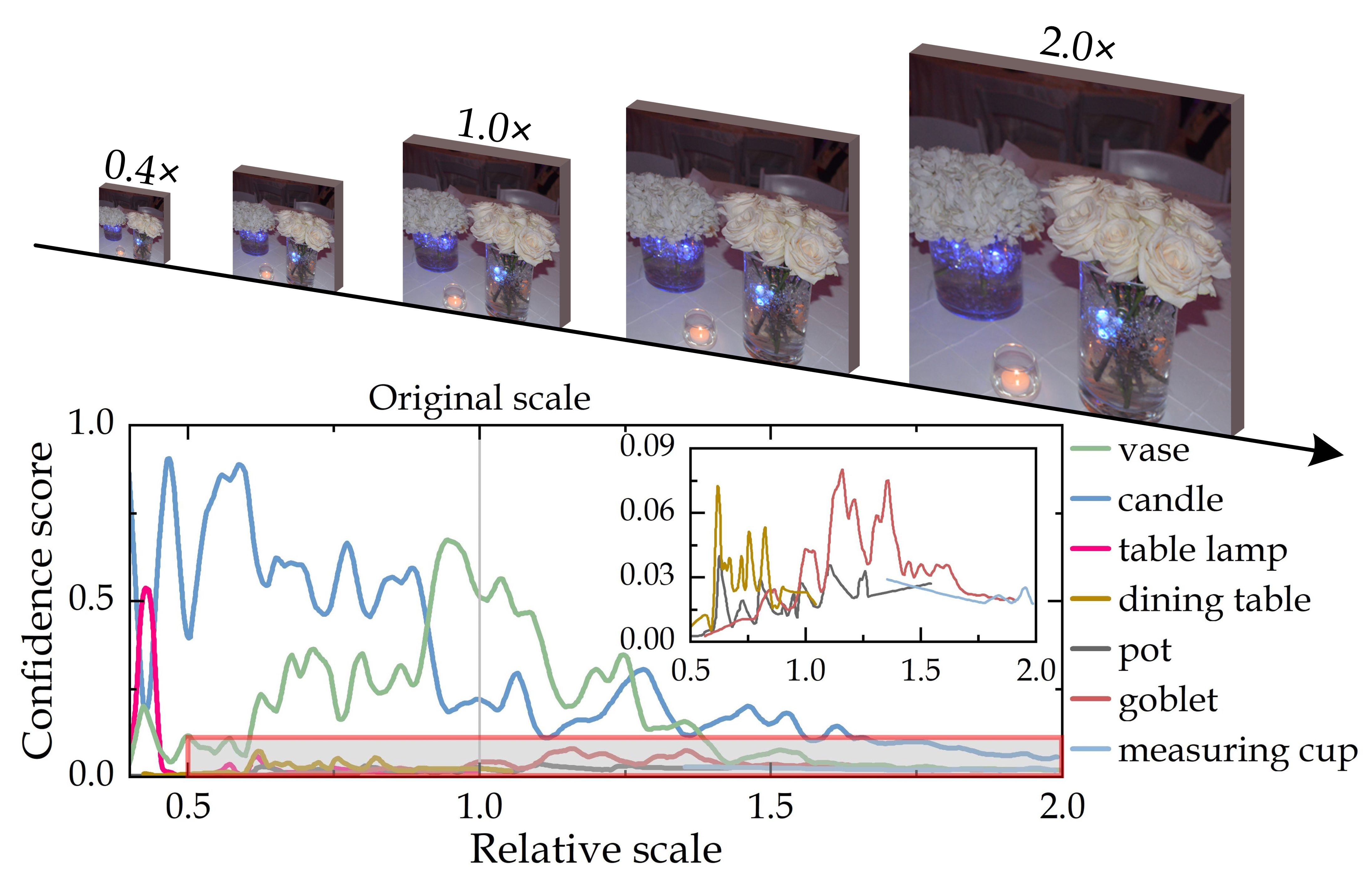}
		\vspace{-5mm}
		\caption{Focus of ResNet50 varies significantly across scales.}
		\label{confidence_score_illu}
		\vspace{-5mm}
	\end{figure}
	
	Simple image scaling, with its inherent simplicity and efficiency, will be the focus of our further investigation, as motivated by the following reasons. A scene often contains multiple objects, and the human visual system may focus on different objects at varying scales. For neural networks, which are usually trained with single-label supervision \cite{resnet2016kh,densenet2017huang,russakovsky2015imagenet}, also attach importance to certain, usually a single predominant object, and overlook the others at a specific scale \cite{Yun_2021_CVPR}. Therefore, scaling is an intuitive and natural way to activate the intrinsic but overlooked semantics for our goal by forcing the mismatch between images and the fixed receptive field of feature extraction backbones. A quantitative instance (GT: \texttt{vase}) can be found in Fig. \ref{confidence_score_illu}. In the context of varying the scale of an image, the model's confidence across different classes will significantly alter. At larger scales, the dominance of a single class is markedly diminished. Conversely, the confidence distribution becomes more sensitive when the scale is smaller. It's worth noting that the semantics themselves can also change across different scales, potentially not corresponding to a label of a specific object that obviously exists in the image but acting instead as cues supporting the judgment of certain classes.
	
\begin{figure}[tbp]
		\includegraphics[width=1\linewidth]{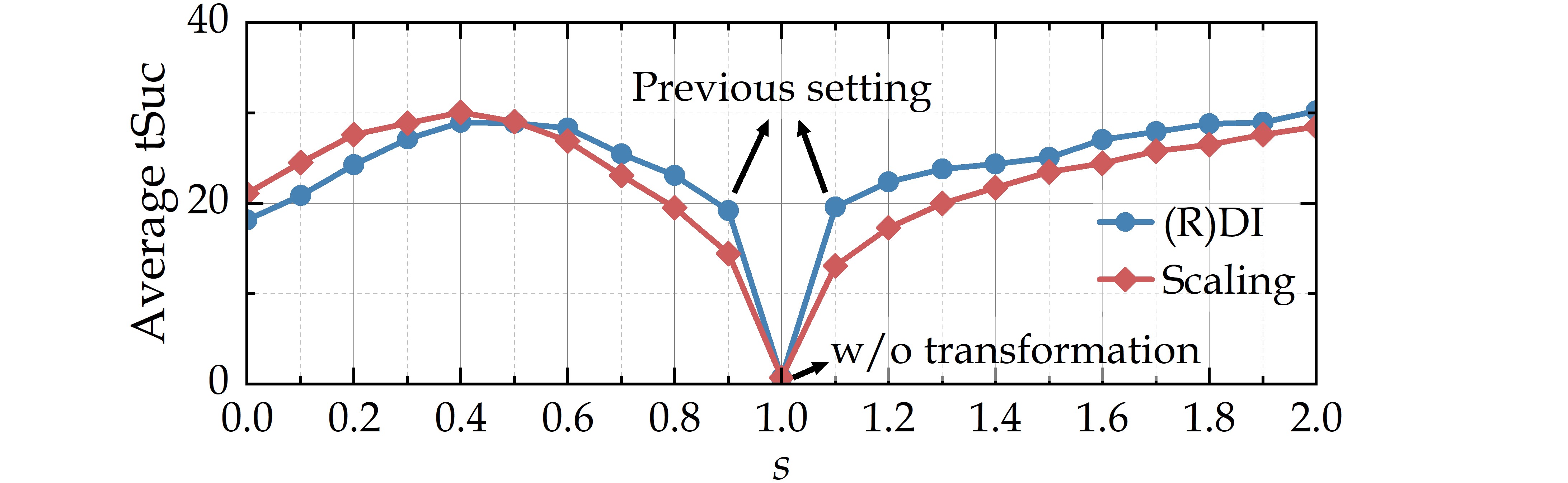}
		\vspace{-5mm}
		\caption{More intense transformation beyond the previous setting further boosts targeted transferability, which aligns with our self-universal hypothesis. Simple image scaling can outperform advanced methods engineered based on it. 
		}
		\label{casestudy}
		\vspace{-6mm}
	\end{figure}

	\textbf{Simple Scaling \textit{vs}. (R)DI}. Scaling is a core operation of (R)DI methods, which have already shown effectiveness for TTA. With a probability $p$, DI \cite{inputdiversity2019xie} dynamically upscales the input to a random larger scale $s^{\prime}H \times s^{\prime}W, s^{\prime}\sim\mathcal{U}(1,s)$, followed by random padding to $sH \times sW$. RDI \cite{RDI} further downscales the image back to the original dimensions, resulting in a smaller-scale version of $\bm{x}$. We conduct experiments following the setting of \cite{logit} to isolate the impact of the random padding operation in (R)DI methods and validate our motivation. The results are presented in Fig. \ref{casestudy}, where \textit{Scaling} only involves scaling images, without the random padding used in (R)DI methods, yet maintaining the same scale range $s$ and probability ($p=0.7$). Results indicate that a broader range of scale variations can significantly enhance attack performance, validating the self-universal hypothesis. Meanwhile, the results highlight the effectiveness of simple scaling since it can achieve better results than (R)DI, showcasing its potential to facilitate the self-universal TTA.
	
	\textbf{Scalability, Sufficiency, and Necessity of Scaling for Self-Universal TTA}. We further extend our investigation to study the characteristics of simple scaling, specifically focusing on its scalability, sufficiency, and necessity for TTA. Building upon the preceding analysis, a transformation that facilitates the exploitation of general \textit{overlooked semantics} should exhibit enhanced transferability, regardless of its similarity or orthogonality to scaling. Therefore, we devise the following experiments for our purpose. 
	
	Assuming an index applicable to any transformation, denoted as $I(\mathcal{T})$, which quantifies the extent to which semantics is leveraged across different \textit{scales} by $\mathcal{T}$-TMI. Simultaneously, using the average tSuc as a substitute for targeted transferability. Therefore, if an increase in $I(\textit{Scaling})$ by expanding the range of scale variations leads to an observed improvement in average tSuc, it can empirically demonstrate the sufficiency of scaling for targeted transferability. Conversely, if an increase in average tSuc caused by scaling-irrelevant transformations leads to an increase in $I(\mathcal{T})$, scaling can be considered empirically necessary. 
	
	The foundation of our design for the $I$ index stems from two natural observations: \textit{1)} categories supported by inherent semantics tend to rank higher in predictions $\bm{p}$; and  \textit{2)} the more intensively the exploitation (attack) on a category, the lower its ranking ought to be after the attack. Therefore, we first define
	\vspace{-2mm}
	\begin{equation}
		\begin{aligned}
			\operatorname{AvgRank}_\mathcal{T}&(f, s, k)=\\
			&\mathbb{E}_{ \bm{x}\sim\mathcal{X}}\frac{1}{k} \sum_{i=1}^{k} \operatorname{rank}(f([\bm{x}_{\mathcal{T}}^{adv}]_{s}), C(f([\bm{x}]_{s}), i)),
		\end{aligned}
		\vspace{-2mm}
	\end{equation}
	where $[\cdot]_{s}$ represents scaling the image to a relative scale $s$,
	$\bm{x}_{\mathcal{T}}^{adv}$ is the AE of $\bm{x}$ generated by $\mathcal{T}$-TMI, and $\operatorname{rank}(\bm{p}, y)$ returns the ranking position of label $y$ within $\bm{p}$. By this design, $\operatorname{AvgRank}_\mathcal{T}(f, s, k)$ denotes the average rank of the original top-$k$ predicted labels for the AE (scaled to scale $s$) crafted by TMI with transformation $\mathcal{T}$. Further, we average $\operatorname{AvgRank}_\mathcal{T}$ on a series of relative scales $\mathcal{S}$:
	\vspace{-2mm}
	\begin{equation}
		\centering
		\hspace{-2.5mm}I(\mathcal{T},k) = \mathbb{E}_{s\sim\mathcal{S}} (\operatorname{AvgRank}_\mathcal{T}(f, s, k)+\operatorname{AvgRank}_\mathcal{T}(f, \frac{1}{s}, k)).
		\vspace{-2mm}
	\end{equation}
	The larger value of $I(\mathcal{T},k)$, \ie the lower ranking of the original top-$k$ predictions, indicates the stronger scaling-induced semantics exploitation of $\mathcal{T}$-TMI. This design exclusively involves the quantities associated with the scaling transformation, thereby minimally reducing the possible interactions with scaling-irrelevant transformations.
	
\begin{figure}[tbp]
		\includegraphics[width=1\linewidth]{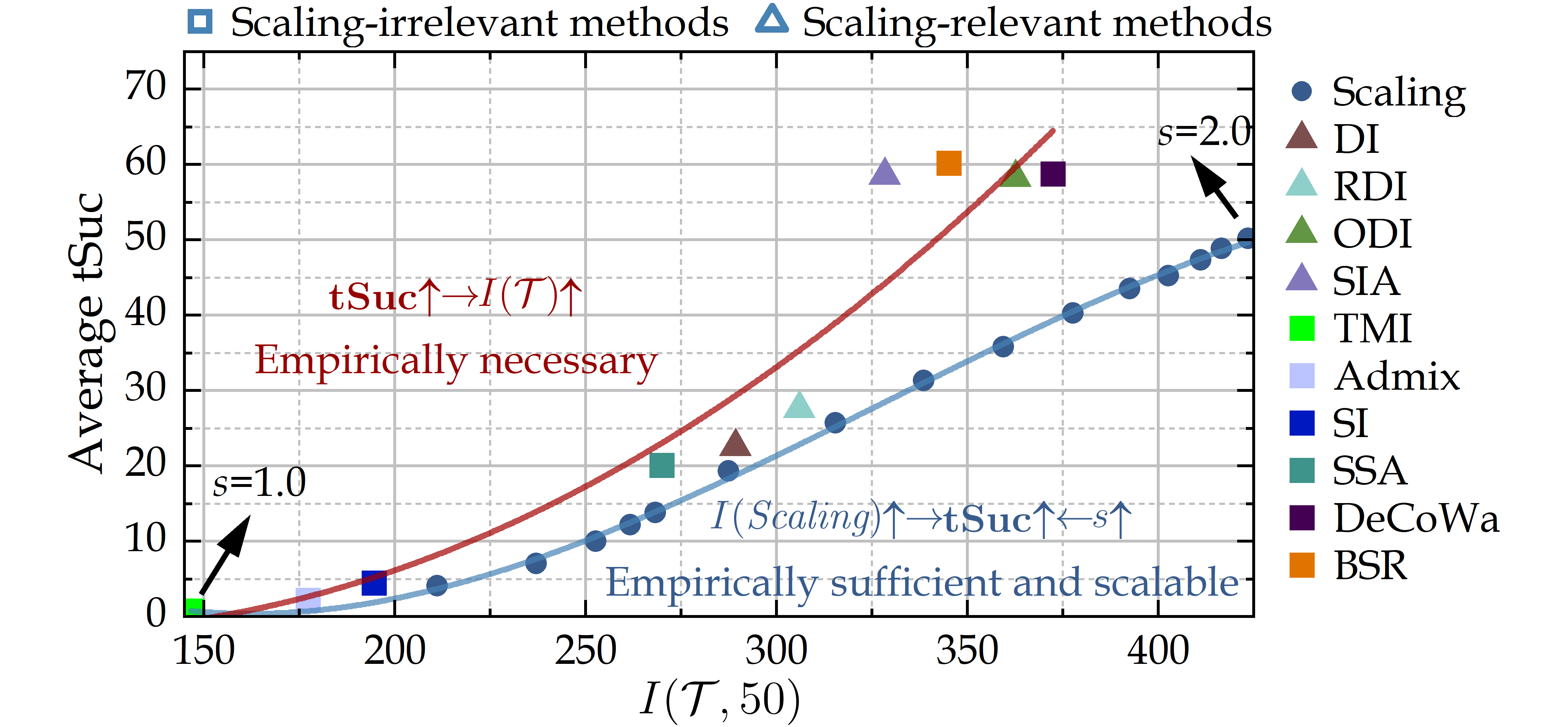}
		\vspace{-5mm}
		\caption{The increase in $I(\textit{Scaling})$ with the expansion of the scaling range demonstrates the rationality of the design of $I$. Further, the positive correlation between $I(\textit{Scaling})$ and the corresponding average tSuc empirically proves the scalability and sufficiency of scaling for targeted transferability. Meanwhile, the positive correlation between the $I(\mathcal{T})$ of scaling-irrelevant transformations and their average tSuc demonstrates the empirical necessity of semantic utilization at varying scales for self-universal TTA.}
		\label{rank_plot}
		\vspace{-5mm}
	\end{figure}

	We include various types of representative methods that are widely studied (as shown in Fig. \ref{existing}). Herein, the simple scaling is extended to cover both upscaling and downscaling, \ie sample  $s^{\prime}\sim\mathcal{U}(\frac{1}{s}, s), s>1$ at each iteration. We depict the relationships between $I(\mathcal{T},50)$ and the corresponding average tSuc against 14 black-box models in Fig. \ref{rank_plot}, where $\mathcal{S}$ is set to the arithmetic sequence ranging from 1.2 to 2.0 with a common difference of 0.2. Other implementation details are deferred to Section \ref{expdetails}.
	
	The metric $I(\textit{Scaling})$ is observed to have a positive correlation with the scaling range $s$ and the corresponding average tSuc, as fitted by the blue curve. This indicates the rationale behind our index $I$ design and the empirical scalability and sufficiency of simple scaling in self-universal TTA. Additionally, results from other types of transformations, including SI, Admix, SSA, DeCoWa, and BSR, despite not involving scaling transformations, exhibit a positive correlation with $I(\mathcal{T})$ (fitted by the red curve), suggesting the empirical necessity of scaling for targeted transferability. It also implies the need to promote the integration of scaling further with the design of other transformations to pursue complementary gains, bridging the gap between the two curves. It is important to acknowledge that this experiment is incomplete as we cannot exhaustively consider all possible transformations and their combinations. Nonetheless, we believe that insights provided by these published and extensively researched methods are highly valuable and persuasive.
	
	\vspace{-4mm}
	\subsection{S$^4$ST}
	\vspace{-1mm}
	Building upon the insights from previous analyses, we refine the simple scaling and enhance it with current advances, culminating in the S$^4$ST transformation. The S$^4$ST comprises a basic operation and two orthogonal enhancements, presented in the following sequence.
	\vspace{-1mm}
	\begin{itemize}
		\item S$^4$ST-\textit{Base}. First, we modify the simple scaling so that the dimensions of the transformed image are consistent with the input one, which facilitates the subsequent extension. Formally, with probability $p_\textit{s}$, S$^4$ST-\textit{Base} samples a relative scale from $s^{\prime}\sim\mathcal{U}(\frac{1}{s},s), s>1$. It downscales the image to $s^{\prime}H\times s^{\prime}W$ and pads it back to the original dimension when $s^{\prime}<1$. For $s^{\prime}>1$, a region of size $\frac{1}{s^{\prime}}H\times \frac{1}{s^{\prime}}W$ is randomly selected from the image and upscaled to the original dimension.
		\item S$^4$ST-\textit{Aug}. We are encouraged to integrate other types of transformations into S$^4$ST-\textit{Base} for complementary gains, according to the observation from Fig. \ref{rank_plot}. With probability $p_{\textit{aug}}$, S$^4$ST-\textit{Aug} pre-applies a random one from a pool of usual transformations \cite{SIA}, with equal probability, to the whole input image. The pool includes image shift and flip in both vertical and horizontal directions, clockwise rotation, linear scale, additive noise, low-pass filtering in the DCT domain, and dropout operations. 
		\item S$^4$ST-\textit{Block}. Recent studies have demonstrated the effectiveness of block-wise operation in boosting transferability \cite{SIA,BSR}. We consider boosting diversity via S$^4$ST-\textit{Block} further, randomly splitting the image into $m$ blocks and applying S$^4$ST-\textit{Base} to each.
	\end{itemize}
	\vspace{-1.5mm}
	We term the integration of the above three components as the S$^4$ST transformation, and illustrations are depicted in Fig. \ref{s4st}.
	
	\textbf{Relationships with SIA and BSR.} The proposed S$^4$ST-\textit{Base} is derived from our self-universal motivation and the analyses upon scaling. The S$^4$ST-\textit{Aug} and S$^4$ST-\textit{Block} strategies are largely inspired by SIA \cite{SIA} and BSR \cite{BSR}, yet exhibit significant differences from them. First, whereas SIA applies random transformations to each image block, our method applies a single random transformation to the entire image, with a focus on block-wise scaling. Second, we do not employ the random shuffling of image blocks nor the small-range continuous rotation transformations proposed in BSR.

	\vspace{-5mm}
	\section{Experiments}
	\vspace{-1mm}
	\subsection{Setup}\label{expdetails}
	\vspace{-1mm}
	
	\textbf{Dataset}. We perform experiments on the ImageNet-Compatible dataset\footnote{\url{https://github.com/cleverhans-lab/cleverhans/tree/master/cleverhans_v3.1.0/examples/nips17_adversarial_competition/dataset}}, which was widely used to assess the performance of TTA \cite{logit,SelfU,ODI}. It was released for the NIPS 2017 adversarial competition, comprises 1,000 299$\times$299 images, and each has a targeted label. We resize the images to 224$\times$224 for evaluation.
	
	\textbf{Models}. According to previous research, we mainly generate AEs based on ResNet-50 \cite{resnet2016kh} and transfer them against dissimilar victim models, including CNNs—\{MobileNet-v2 (MNv2) \cite{Sandler_2018_CVPR}, EfficientNet-b0 (EN) \cite{efficientnet}, ConvNeXt-small (CNX) \cite{liu2022convnet}, Inception-v3 (INv3) \cite{szegedy2016rethinking}, Inception-v4 (INv4) \cite{szegedy2017inception}, Inception ResNet-v2 (IRv2), and Xception (Xcep) \cite{chollet2017xception}\} and vision transformers—\{ViT \cite{dosovitskiy2020image}, SwinT \cite{Liu_2021_ICCV}, MaxViT \cite{tu2022maxvit}, Twins \cite{chu2021twins}, PiT \cite{heo2021rethinking}, TNT \cite{NEURIPS2021_854d9fca}, and DeiT \cite{pmlr-v139-touvron21a}\}. We collect their pre-trained weights from \textit{torchvision}\footnote{\url{https://github.com/pytorch/vision/tree/master/torchvision/models}} and \textit{timm} \cite{rw2019timm} libraries.
	
	\textbf{Metrics}. The transferability of an attack algorithm is measured by the average tSuc, \ie the percentage of success transfer over the dataset from the surrogate model to the victim models. We also compare the time consumption of different transformations in crafting AEs. All experiments were conducted on a single NVIDIA GeForce RTX 4090 GPU with Pytorch \cite{NEURIPS2019_bdbca288} implementation.
	
	\textbf{Baseline}. We use the TMI attack as a baseline to compare different methods. The step size and perturbation budget are set to 2/255 and 16/255, respectively. The decay factor for momentum accumulation is set to 1 \cite{momentum2018dong}, and we use a 5$\times$5 Gaussian kernel \cite{evading2019dong} for TI. We use $\mathcal{L}_{\text{NCE}}^{\text{MLC}}$ \cite{marginangleT} and set $T=900$. 
	
\begin{figure}[tbp]
		\includegraphics[width=1\linewidth]{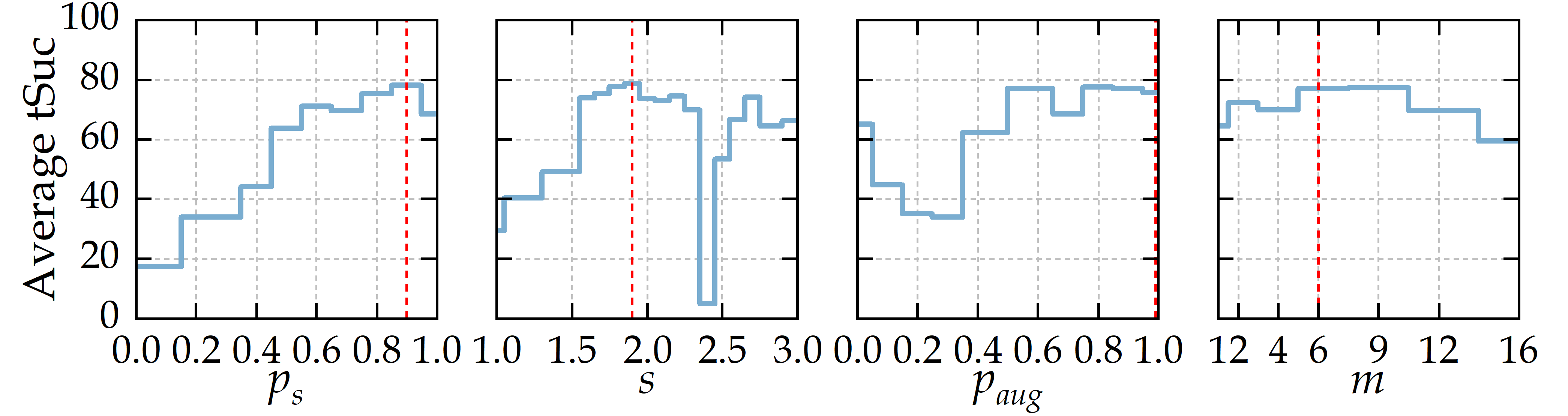}
		\vspace{-6mm}
		\caption{Partial dependence of the hyper-parameters resulted from the Bayes optimization over 100 calls.}
		\label{partialdependency}
		\vspace{-6mm}
	\end{figure}

	\textbf{Hyper-Parameter Settings.} We study the effects of hyper-parameters of our methods prior to the main experiments, including $p_{\textit{s}}$, $s$, $p_{\textit{aug}}$, and $m$. Two considerations merit particular attention. Initially, it should be noted that while stronger diversity, as per our analysis, should result in better targeted transferability; it could also lead to an increased number of iterations required for convergence. We can only search for the optimal parameter combination with a specific number of iterations; also, for this reason, optimizing parameters for a single component might lead to sub-optimal results when combined. Therefore, we employ Bayesian optimization to perform a joint search. The search is performed on ResNet-50 \cite{resnet2016kh} over 100 images using the \textit{gp\_minimize} function from \textit{scikit-optimize}\footnote{\url{https://scikit-optimize.github.io/stable/}}, which involved 10 random starts and a total of 100 calls. The resulted parameter combination is $p_{\textit{s}}=0.9, s=1.9, p_{\textit{aug}}=1.0, m=6$, where $m=6$ indicates partitioning the image into 2 and 3 segments along the respective coordinate axes (not specifying horizontal or vertical orientation in order). The partial dependence of hyper-parameters on the attack performance is reported in Fig. \ref{partialdependency}. For other transformation methods, we follow the implementations in respective papers and the \textit{TransferAttack}\footnote{\url{https://github.com/Trustworthy-AI-Group/TransferAttack}} code base.

	\begin{table*}[tbp]
		\caption{Comparison between input transformations based on ResNet-50 as the surrogate model.}
		\label{rn50_trans}
		\centering
		\vspace{-2mm}
		\resizebox{\linewidth}{!}{
			\begin{tabular}{l|cc|cccccccc|cccccccc}
				\toprule[1pt]
				Attack & tSuc & Time & MNv2 & EN & CNX  & INv3 & INv4 & IRv2 & Xcep & AvgC & ViT  & Swin & MaxViT & Twins & PiT & TNT & DeiT & AvgT  \\ \midrule
				TMI \cite{momentum2018dong,evading2019dong} & 0.8 & 0.28 & 2.0 & 1.5 & 2.4 & 0.6 & 0.7 & 0.7 & 0.9 & 1.3 &  0.2 & 0.9 & 0.4 & 0.1 & 0.1 & 0.2 & 0.2 & 0.3 \\ 
				+Admix \cite{admix} & 2.1 & 4.99 & 5.9 & 4.7 & 6.3 & 0.9 & 1.6 & 2.4 & 2.5 & 3.5 & 0.3 & 2.0 & 1.0 & 0.5 & 1.1 & 0.6 & 0.2 & 0.8 \\  
				+SI \cite{nestrov2019lin} & 4.4 & 4.24 & 8.5 & 9.6 & 8.4 & 5.7 & 5.3 & 6.3 & 7.0 & 7.3 &  1.1 & 2.4 & 1.5 & 1.2 & 2.5 & 1.3 & 0.6 & 1.5 \\ 
				+SSA \cite{SSA} & 19.8 & 0.41 & 35.0 & 32.0 & 35.1 & 22.5 & 30.4 & 35.1 & 33.0 & 31.9 &  4.8 & 11.1 & 7.1 & 7.4 & 10.8 & 7.2 & 6.0 & 7.8 \\ 
				+DI \cite{inputdiversity2019xie} & 22.4 & 0.29 & 29.6 & 33.7 & 48.8 & 15.4 & 35.4 & 38.1 & 28.3 & 32.8 &  5.9 & 17.1 & 10.8 & 11.1 & 19.1 & 11.6 & 8.3 & 12.0 \\ 
				+RDI \cite{RDI} & 27.3 & 0.29 & 37.1 & 42.9 & 47.4 & 25.8 & 39.8 & 45.6 & 36.9 & 39.4 & 11.0 & 21.7 & 13.2 & 13.6 & 22.3 & 15.0 & 10.2 & 15.3 \\ 
				+ODI \cite{ODI} & 58.1 & 4.08 & 73.2 & 77.1 & 75.3 & \textbf{64.3} & 75.7 & 77.0 & 73.2 & 73.7 & 37.6 & 54.1 & 37.9 & 38.4 & 54.5 & 43.0 & 31.7 & 42.5 \\ 
				+SIA \cite{SIA} & 58.3 & 1.05 & \textbf{82.5} & \textbf{81.3} & \textbf{84.3} & 51.1 & 74.2 & 76.2 & 72.0 & 74.5 & 30.8 & \textbf{58.5} & \textbf{42.6} & 39.3 & 53.2 & 40.7 & 29.8 & 42.1 \\
				+DeCoWA \cite{decowa} &  58.4 & 1.45 & 74.9 & 78.0 & 77.4 & 56.7 & 74.4 & 77.7 & 73.7 & 73.3 & \textbf{40.6} & 58.4 & 36.1 & 36.6 & 55.1 & 44.1 & 33.3 & \textbf{43.5} \\    
				+BSR \cite{BSR} & 60.0 & 1.14 & 81.8 & 80.3 & 81.4 & 62.6 & \textbf{79.3} & \textbf{78.1} & \textbf{76.6} & \textbf{77.2} & 32.2 & 56.8 & 38.4 & \textbf{39.6} & \textbf{55.7} & \textbf{44.2} & \textbf{33.5} & 42.9 \\ 
				\midrule
				+Scaling & 54.7 & 0.31 &  58.6 & 68.7 & 88.9 & 43.4 & 74.2 & 73.5 & 73.0 & 68.6 &  29.7 & 56.2 & 39.8 & 42.6 & 45.1 & 40.7 & 31.4 & 40.8 \\  
				+S$^4$ST-\textit{Base} & 59.9 & 0.32 &  69.7 & 76.6 & 82.1 & 59.8 & 81.9 & 79.1 & 77.9 & 75.3 &  34.9 & 59.4 & 43.4 & 44.3 & 54.2 & 43.8 & 31.6 & 44.5 \\  
				+S$^4$ST-\textit{Aug}   & 74.5
				& 0.34 &  87.6 & 89.7 & 89.8 & 79.5 & \textbf{90.4} & 88.2 & 89.5 & 87.8 & 51.5 & 74.6 & 57.9 & 62.8 & 70.7 & 63.4 & 46.9 & 61.1 \\   
				+S$^4$ST-\textit{Block} & 75.0  & 0.39 & 89.7 & 90.2 & \textbf{92.7} & 70.9 & 88.9 & 87.0 & 87.9 & 86.8 & 51.5 & 79.2 & 63.4 & 64.2 & 71.1 & 64.7 & 49.2 & 63.3   \\ 
				+S$^4$ST & \textbf{79.8}
				& 0.41 &  \textbf{92.2} & \textbf{91.9} & 92.0 & \textbf{80.2} & 90.3 & \textbf{88.3} & \textbf{89.7} & \textbf{89.2} &  \textbf{61.2} & \textbf{82.5} & \textbf{70.1} & \textbf{70.0} & \textbf{77.6} & \textbf{72.0} & \textbf{58.9} & \textbf{70.3} \\ 
				\bottomrule[1pt]
		\end{tabular}}
		\\
		\vspace{1mm}
		
		\leftline{The best results of compared and our methods are both in bold.}
		\vspace{-2mm}
	\end{table*}
	
	\begin{table*}[tbp]
		\caption{Results of attacking robust models (tSuc/uSuc) and ensemble-based attacks.}
		\vspace{-2mm}
		\label{secured_eval}
		\centering
		\resizebox{\linewidth}{!}{
			\begin{tabular}{l|ccccccccccc|c|c}
				\toprule[1pt]
				\multirow{3}{*}{Attack} & \multirow{3}{*}{{Augmix \cite{hendrycks2019augmix}}} & \multicolumn{2}{c}{Stylized \cite{geirhos2018imagenet}} & \multicolumn{4}{c}{Adversarial training (AT) \cite{madry2018towards}}  & \multicolumn{3}{c}{Ensemble AT \cite{tramer2018ensemble}}  & \multirow{3}{*}{Avg.} & \multirow{3}{*}{AvgC} &  \multirow{3}{*}{AvgT}\\  \cmidrule(lr){3-4}  \cmidrule(lr){5-8}  \cmidrule(lr){9-11}  
				
				&    & SIN & SIN-IN &$\ell_{2}=.1$ & $\ell_{2}=.5$ & $\ell_{\infty}=.5$ & $\ell_{\infty}=1.$ & INv3$_{ens3}$ & INv3$_{ens4}$ & IRv2$_{ens}$ &  &  &  \\ \midrule
				
				\multicolumn{14}{c}{Surrogate: ResNet-50} \\ \midrule
				TMI \cite{momentum2018dong,evading2019dong} & 3.4/55.8 & 0.5/51.6 & 33.7/83.4 & 0.2/42.4 & 0.0/28.6 & 0.0/24.3 & 0.0/19.1 & 0.0/17.8 & 0.0/17.9 & 0.0/12.4 & 3.8/35.3 & 1.3 & 0.3 \\ 
				+ODI \cite{ODI} & 78.5/94.6 & 45.2/87.6 & 96.5/99.4 & 65.6/88.8 & 3.5/44.5 & 6.4/46.2 & 0.6/28.2 & 14.4/51.4 & 5.8/40.1 & 3.2/27.6 & 32.0/60.8 & 73.7 & 42.5 \\
				+SIA \cite{SIA} & 89.1/96.7 & 39.8/84.6 & \textbf{99.6}/\textbf{100.0} & 64.1/87.5 & 1.5/41.5 & 3.7/40.2 & 0.1/27.2 & 5.7/37.7 & 1.7/32.5 & 0.6/19.9 & 30.6/56.8 & 74.5 & 42.1 \\
				+DeCoWA \cite{decowa} & 82.8/95.7 & 45.2/87.9 & 97.7/99.8 & 72.2/91.7 & 4.0/48.5 & 8.2/49.0 & 0.7/30.2 & 16.8/58.2 & 8.8/49.0 & 4.0/34.1 & 34.0/64.4 & 73.3 & 43.5 \\
				+BSR \cite{BSR} & 86.5/96.1 & 48.0/87.0 & 98.9/99.9 & 69.6/89.9 & 2.8/44.9 & 5.5/46.9 & 0.3/25.7 & 11.0/48.7 & 3.8/38.5 & 2.1/27.6 & 32.9/60.5 & 77.2 & 42.9 \\
				+S$^4$ST (Ours)  & \textbf{92.9}/\textbf{98.2} & \textbf{56.0}/\textbf{89.6} & 98.5/99.8 & \textbf{83.9}/\textbf{95.0} & \textbf{9.9}/\textbf{52.9} & \textbf{18.1}/\textbf{55.4} & \textbf{1.3}/\textbf{30.6} & \textbf{34.5}/\textbf{65.6} & \textbf{15.7}/\textbf{51.2} & \textbf{9.8}/\textbf{38.3} & \textbf{42.1}/\textbf{67.7} & \textbf{89.2} & \textbf{70.3} \\
				\midrule
				\multicolumn{14}{c}{Surrogate: ResNet-50 + ResNet-152 + DenseNet-121 + VGG16-bn} \\ \midrule
				TMI$_{\textit{Ens}}$ & 28.0/69.1 & 5.0/56.6 & 67.8/89.5 & 5.4/49.4 & 0.2/28.6 & 0.1/26.2 & 0.0/21.0 & 0.0/19.7 & 0.0/19.4 & 0.0/12.2 & 10.7/39.2 & 23.0 & 9.7 \\ 
				+ODI \cite{ODI}  & 93.0/97.4 & 61.8/88.6 & 98.9/99.8 & 79.7/92.7 & 5.6/48.7 & 10.0/49.8 & 0.9/30.1 & 20.8/55.7 & 10.1/42.9 & 5.6/30.5 & 38.6/63.6 & 91.8 & 73.8 \\
				+SIA \cite{SIA}  & 98.4/99.5 & 68.5/92.5 & \textbf{99.8}/\textbf{99.9} & 90.2/97.1 & 6.9/50.4 & 13.0/52.8 & 1.0/30.8 & 23.5/57.4 & 9.9/43.1 & 5.4/30.2 & 41.7/65.4 & 95.9 & 83.9 \\
				+DeCoWA \cite{decowa} & 97.7/99.0 & 77.0/\textbf{96.2} & 99.4/\textbf{99.9} & 93.4/98.0 & 13.4/57.7 & 21.7/59.8 & 2.4/35.4 & 51.9/79.7 & 32.5/65.6 & 17.7/49.0 & 50.7/74.0 & 95.2 & 82.2 \\
				+BSR \cite{BSR} &  98.1/99.6 & 73.0/94.2 & \textbf{99.8}/\textbf{99.9} & 89.8/96.8 & 7.3/52.0 & 15.4/55.9 & 1.0/31.3 & 35.0/67.1 & 14.0/51.3 & 8.4/37.1 & 44.2/68.5 & 95.4 & 78.7 \\
				+S$^4$ST (Ours)  & \textbf{99.3}/\textbf{99.8} & \textbf{78.0}/95.3 & \textbf{99.8}/\textbf{99.9} & \textbf{95.6}/\textbf{98.8} & \textbf{23.8}/\textbf{61.1} & \textbf{33.7}/\textbf{66.1} & \textbf{3.5}/\textbf{35.6} & \textbf{65.2}/\textbf{84.2} & \textbf{39.3}/\textbf{68.8} & \textbf{25.6}/\textbf{53.6} & \textbf{56.4}/\textbf{76.3} & \textbf{98.1} & \textbf{91.2} \\
				\bottomrule[1pt]
		\end{tabular}}
		\\
		\vspace{1mm}
		\leftline{The best results are in bold. uSuc is the ratio (in \%) of the false predictions caused by the targeted AE by the original correct predictions.}
		\vspace{-5mm}
	\end{table*}
	
	\vspace{-4mm}
	\subsection{Comparison with Existing Transformations}\label{comptrans}
	\vspace{-1mm}
	\textbf{Evaluation on Single Model}. We first compare S$^4$ST with other input transformation methods. The transformed copies per iteration for SSA, SIA, BSR, and DeCoWa are limited to 1 since this will yield better results than multiple copies but with fewer iterations. We also set $m_{1}=1, m_{2}=5$ for Admix to ensure a fair comparison with SI. Results are listed in Table \ref{rn50_trans}. Relative to the baseline, S$^4$ST transformations incur limited additional computational overhead yet yield very compelling results compared to the current most powerful and time-consuming competitors. Even S$^4$ST-\textit{Base} achieves 59.9\% average tSuc, and the variations further significantly boost the transferability, up to an average tSuc of 79.8\% against all evaluated CNNs and transformers. We extend the comparison to alternative surrogate models without further fine-tuning the hyper-parameters, and our methods consistently outperform the others, as shown in Table \ref{alter_trans}. 
	
	\begin{table}[tbp]
		\caption{Results by alternative surrogate models.}
		\label{alter_trans}
		\centering
		\vspace{-2mm}
		\resizebox{0.9\linewidth}{!}{
			\begin{tabular}{l|cc|cc|cc}
				\toprule[1pt]
				Surrogate & \multicolumn{2}{c|}{DenseNet-121 \cite{densenet2017huang}} & \multicolumn{2}{c|}{VGG16-bn \cite{vgg2015sk}} & \multicolumn{2}{c}{ViT \cite{dosovitskiy2020image}} \\ \hline
				Attack & tSuc & Time & tSuc & Time & tSuc & Time \\ \midrule
				TMI \cite{momentum2018dong,evading2019dong} & 1.1 & 0.86 & 0.8 & 0.11 & 0.0 & 0.39 \\ 
				+ODI \cite{ODI} & 46.2 & 4.66 & 38.3 & 4.12 & 31.1 & 6.17 \\ 
				+SIA \cite{SIA} & \textbf{49.9} & 1.32 & 33.7 & 1.48 & 22.1 & 3.05 \\ 
				+DeCoWA \cite{decowa} & 43.6 & 2.15 & \textbf{46.5} & 2.17 & 14.6 & 4.78 \\ 
				+BSR \cite{BSR} & 47.9 & 1.25 & 32.1 & 2.03 & \textbf{27.6} & 4.35 \\ 
				\midrule
				+Scaling & 36.4 & 1.21 & 31.1 & 0.14 & - & - \\
				+S$^4$ST-\textit{Base} & 43.9 & 0.92 & 36.5
				& 0.14 & 32.3 & 0.47 \\    
				+S$^4$ST-\textit{Aug}  & 61.0
				& 0.94 & 57.5
				& 0.15 & 46.0 & 0.49 \\  
				+S$^4$ST-\textit{Block} & 64.3
				& 0.98 & 46.7
				& 0.20 & 52.0 & 0.53 \\ 
				+S$^4$ST & \textbf{71.7}
				& 1.00 & \textbf{63.0}
				& 0.21 & \textbf{60.6} & 0.55 \\ 
				\bottomrule[1pt]
			\end{tabular}
		}
		\\
		\vspace{1mm}
		\leftline{The best results of compared and our methods are both in bold.}
		\vspace{-8mm}
	\end{table}
	
	\begin{table*}[htb]
		\caption{Comparison with resource-intensive methods under the 10-Targets (all-source) setting.}
		\vspace{-2mm}
		\label{agnostic_comp}
		\centering
		\resizebox{\linewidth}{!}{
			\begin{tabular}{l|cccccc|cccccc|cccccc}
				\toprule[1.pt]
				\multirow{3}{*}{Attack}	& \multicolumn{6}{c}{CNNs} & \multicolumn{6}{c}{Vision transformers} &\multicolumn{6}{c}{Robust mechanisms} \\  \cmidrule(lr){2-7}  \cmidrule(lr){8-13}  \cmidrule(lr){14-19} 
				& CNX & INv3 & INv4 & IRv2 & Xcep & Avg. & ViT & Swin  & PiT & TNT & DeiT  & Avg. & Augmix & SIN & SIN-IN & AT$_{\ell_{2}=.5}$ & AT$_{\ell_{\infty}=.5}$  & Avg.  \\ \midrule
				\multicolumn{19}{c}{Resource-intensive TTAs} \\ \midrule
				DFA \cite{domainfeatureuap} &  25.3 & 1.6 & 15.0 & 16.3 & 13.8 & 14.4 &  0.7 & 3.1 & 1.6 & 1.7 & 3.9 & 2.2 &  19.9 & 7.8 & 56.4 & 0.3 & 0.3 & 16.9 \\
				BSR+DFA \cite{BSR,domainfeatureuap} & 46.0 & 16.2 & 37.8 & 37.5 & 35.1 & 34.5 &  4.5 & 9.8 & 6.9 & 7.3 & 7.0 & 7.1 & 33.0 & 17.2 & 49.9 & 0.6 & 0.9 & 20.3  \\
				TTP \cite{naseer2021generating} & 54.7 & 38.4 & 56.6 & 57.9 & 60.6 & 53.6 &  30.0 & 35.6 & 24.4 & 28.6 & 11.4 & 26.2 & 68.7 & 27.8 & 84.1  &  2.9 & 5.4 & 37.8 \\		
				M3D \cite{zhao2023minimizing} &	\textbf{84.0} & \textbf{65.9} & \textbf{83.5} & \textbf{85.9} & \textbf{84.4} & \textbf{80.7} &  \textbf{60.8} & \textbf{67.7} & \textbf{62.0} & \textbf{62.2} & \textbf{41.0} & \textbf{58.7} & \textbf{88.0} & \textbf{62.9} & \textbf{94.6}  &  \textbf{5.2} & \textbf{9.2} & \textbf{52.0} \\		\midrule
				\multicolumn{19}{c}{Simple TTAs} \\ \midrule
				TMI \cite{momentum2018dong,evading2019dong} & 5.4 & 1.1 & 1.9 & 2.1 & 2.1 & 2.5 & 0.8 & 1.7 & 1.1 & 1.0 & 0.7 & 1.1 & 4.4 & 1.4 & 37.1 & 0.4 & 0.4 & 8.7 \\
                +ODI \cite{ODI} & 74.3 & 62.6 & 73.6 & 75.1 & 70.8 & 71.3 & 37.6 & 53.3 & 55.5 & 42.8 & 32.2 & 44.3 & 78.1 & 45.2 & 96.8 & 6.1 & 9.6 & 47.2 \\
				+SIA \cite{SIA} &  87.5 & 53.9 & 78.5 & 79.9 & 73.6 & 74.7 &  34.6 & 64.2 & 58.1 & 47.8 & 35.6 & 48.1 & 90.6 & 44.1 & \textbf{99.7} & 3.6 & 5.9 & 48.8 \\
				+BSR \cite{BSR} & 83.5 & 62.3 & 79.4 & 80.4 & 73.9 & 75.9 &  35.6 & 59.8 & 60.6 & 48.6 & 34.3 & 47.8 & 88.0 & 52.1 & 98.8 & 5.2 & 8.9 & 50.6 \\
				+S$^4$ST (Ours) &  \textbf{93.9} & \textbf{81.3} & \textbf{91.6} & \textbf{91.5} & \textbf{91.1} & \textbf{89.9} & \textbf{66.2} & \textbf{86.1} & \textbf{81.8} & \textbf{77.1} & \textbf{61.9} & \textbf{74.6} &  \textbf{93.2} & \textbf{60.0} & 98.4 & \textbf{14.7} & \textbf{23.2} & \textbf{57.9} \\			
				\bottomrule[1.pt]
		\end{tabular}}
		\\ \vspace{1mm}
		\leftline{The best results in respective groups are in bold.}
		\vspace{-7mm}
	\end{table*}
	
	Fig. \ref{t-s-plot} further reports the results under different iterations. It can be observed that the simple scaling outperforms almost all competitors when  $T=300$, further corroborating our analysis that simple scaling can efficiently activate the intrinsic multi-class semantics. Additionally, more optimization iterations have shown to be significantly beneficial for stronger methods, especially ours.

	\textbf{Evaluation on Robust Victims}. Table \ref{secured_eval} reports the tSuc and untargeted transfer success rate (uSuc) against robust models. Following the evaluation protocol of previous research \cite{naseer2021generating,zhao2023minimizing}, we include Augmix \cite{hendrycks2019augmix}, stylized training (SIN and SIN-IN) \cite{geirhos2018imagenet}, adversarial training (AT) \cite{madry2018towards,salman2020adversarially}, and Ensemble AT \cite{tramer2018ensemble}. While targeted prediction is hard to trigger at the adversarially trained models with large perturbation budgets, they are also prone to give wrong predictions to the targeted AEs. Our S$^4$ST achieves the best results on both metrics. 
	
	\textbf{Evaluation on Ensemble Models}. The ensemble-based strategy is further leveraged to mount more transferable attacks \cite{momentum2018dong,logit}. We limit the surrogate models to be dissimilar from the victims, and the results are shown in the second block of Table \ref{secured_eval}. Substantial improvements can be observed compared to the single surrogate case, especially a significant enhancement in the attack capabilities against transformer models and adversarially trained models, where the proposed method also achieves the best results.
	
	\vspace{-4mm}
	\subsection{Comparison with Resource-Intensive Methods}
	\vspace{-1mm}
	We compare input transformation-empowered simple attacks with the resource-intensive ones, including generative methods—\{TTP \cite{naseer2021generating} and M3D \cite{zhao2023minimizing}\} and universal attacks—\{Dominant Feature Attack (DFA) \cite{domainfeatureuap} and BSR+DFA\}. It is computationally forbidden to train these methods to attack the massive target labels specified by the ImageNet-Compatible dataset. Therefore, we follow the 10-Targets (all-source) setting \cite{naseer2021generating} that averages tSuc on 10 target labels over the dataset. We use the pre-trained generators provided by the authors for TTP and M3D, which were trained over 50,000 additional images from the ImageNet train set. We also allow 50,000 images from the ImageNet validation set for DFA training. Results are reported in Table \ref{agnostic_comp}. It is observed that our S$^4$ST significantly enhances the baseline attack and exceeds the SOTA resource-intensive method, M3D. This suggests that exploiting the specific knowledge of the target image itself can lead to better targeted transferability than utilizing overall distributional knowledge.
	
	\vspace{-4mm}
	\subsection{Transferring Against Real World API}
	\vspace{-1mm}
	Evaluation against the commercially deployed models in the real world is performed to verify the practicality of our method. We follow the evaluation setting of M3D to perform transfer attacks on 5 labels over 100 images against the Google Cloud Vision API (label detection)\footnote{\url{https://cloud.google.com/vision/docs/labels}}. tSuc here is measured if the target label appears in the returned label list. Semantically similar labels are treated as the same one. We compare S$^4$ST-TMI and M3D as well as provide the results of S$^4$ST-TMI$_{\textit{Ens}}$ in Table \ref{apis}. It shows that the average tSuc of our method is over 22\% compared to M3D, and utilizing ensemble surrogates further achieves an average tSuc up to 70\%. These impressive results indicate our S$^4$ST-empowered attacks are promising in assessing the vulnerability of models with unknown training data distribution and data processing schemes.

	\begin{table}[tbp]
		\centering
		\caption{Results of attacking Google Cloud Vision API.}
		\label{apis}
		\vspace{-2mm}
		\resizebox{1.0\linewidth}{!}{
			\begin{tabular}{cccc|c}
				\toprule[1pt]
				Target class & Similar classes & M3D \cite{zhao2023minimizing} & S$^4$ST-TMI & S$^4$ST-TMI$_{\textit{Ens}}$ \\ \midrule
				Grey-Owl & Owl, Screech owl, great grey owl & 23 & 42 & 55 \\
				Goose & geese and swans & 39 & 63 & 54 \\
				Bulldog & Dog, Companion dog  & 46 & 56 & 88 \\
				Parachute  &  Parachute  & 28 & 70 & 80 \\
				Street sign & Signage, Street sign & 39 & 54 & 73 \\ \cmidrule{3-5}
				\multicolumn{2}{r}{Average tSuc} & 35 & 57 & 70 \\
				\bottomrule[1pt]
		\end{tabular}}
		\vspace{-4mm}
	\end{table}

	\vspace{-4mm}
	\subsection{Visualizations}
\begin{figure}[tbp]
		\includegraphics[width=1\linewidth]{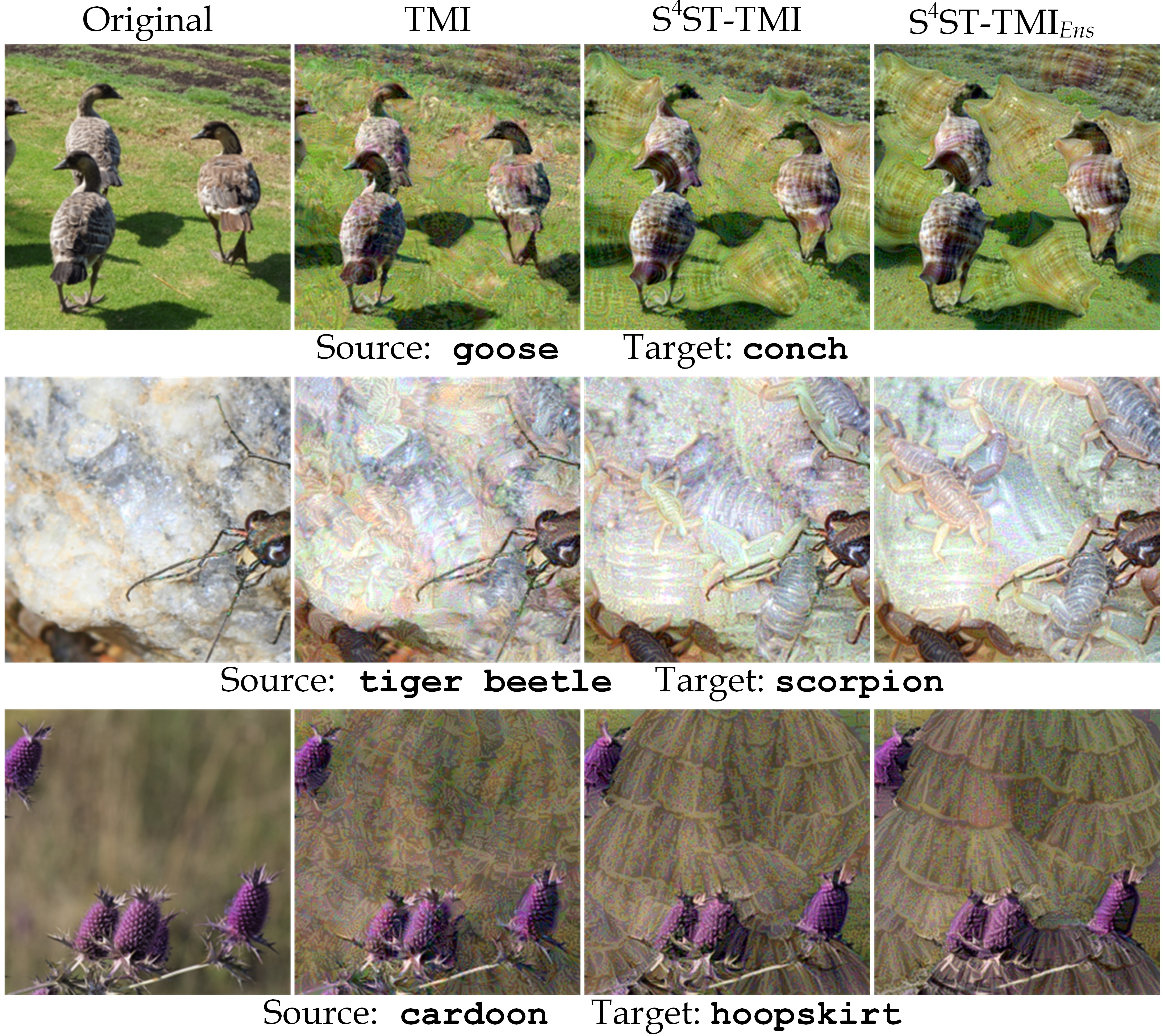}
		\vspace{-5mm}
		\caption{Adversarial examples crafted by different attacks.}
		\label{vis}
		\vspace{-6mm}
	\end{figure}

	\vspace{-1mm}
	Visual instances are presented in Figure \ref{vis}. It reveals that our S$^4$ST can significantly facilitate the generation of the target label's semantics, which can be further advanced by attacking an ensemble of surrogate models. Similar observations can be obtained in resource-intensive attacks \cite{naseer2021generating,domainfeatureuap}, which directly supports our self-universal strategy. However, transformation-empowered simple attacks enjoy greater focus on manipulating the existing content and texture within individual images, thereby resulting in better transferability.
	
	\vspace{-4mm}
	\section{Conclusion}\label{conclusion}
	\vspace{-1mm}
	In the last few years, input transformation techniques have been revealed as key facilitators of targeted attack transferability but have only been lightly explored. In this paper, we propose a self-universal hypothesis to comprehend its role. Further, we are the first to thoroughly investigate the potential of simple image scaling as an input transformation and, based on this, introduce the S$^4$ST, which facilitates simple scaling with extra complementary improvements. Comprehensive experiments have validated the superior effectiveness and efficiency of our proposed approach to enhance the targeted transferability. We hope our insights provide some theoretical and practical perspectives on adversarial transferability and eventually benefit the understanding and trustworthiness of deep neural networks.
	
	\bibliographystyle{IEEEtran}    
	\bibliography{s4st_refs}         

\begin{thebibliography}{10}
\providecommand{\url}[1]{#1}
\csname url@samestyle\endcsname
\providecommand{\newblock}{\relax}
\providecommand{\bibinfo}[2]{#2}
\providecommand{\BIBentrySTDinterwordspacing}{\spaceskip=0pt\relax}
\providecommand{\BIBentryALTinterwordstretchfactor}{4}
\providecommand{\BIBentryALTinterwordspacing}{\spaceskip=\fontdimen2\font plus
\BIBentryALTinterwordstretchfactor\fontdimen3\font minus
  \fontdimen4\font\relax}
\providecommand{\BIBforeignlanguage}[2]{{%
\expandafter\ifx\csname l@#1\endcsname\relax
\typeout{** WARNING: IEEEtran.bst: No hyphenation pattern has been}%
\typeout{** loaded for the language `#1'. Using the pattern for}%
\typeout{** the default language instead.}%
\else
\language=\csname l@#1\endcsname
\fi
#2}}
\providecommand{\BIBdecl}{\relax}
\BIBdecl

\bibitem{szegedy2013intriguing}
C.~Szegedy, W.~Zaremba, I.~Sutskever, J.~Bruna, D.~Erhan, I.~Goodfellow, and
  R.~Fergus, ``Intriguing properties of neural networks,'' in \emph{ICLR},
  2014.

\bibitem{harnessing2015goodfellow}
I.~J. Goodfellow, J.~Shlens, and C.~Szegedy, ``Explaining and harnessing
  adversarial examples,'' in \emph{ICLR}, 2015.

\bibitem{madry2018towards}
A.~Madry, A.~Makelov, L.~Schmidt, D.~Tsipras, and A.~Vladu, ``Towards deep
  learning models resistant to adversarial attacks,'' in \emph{ICLR}, 2018.

\bibitem{tramer2018ensemble}
F.~Tramèr, A.~Kurakin, N.~Papernot, I.~Goodfellow, D.~Boneh, and P.~McDaniel,
  ``Ensemble adversarial training: Attacks and defenses,'' in \emph{ICLR},
  2018.

\bibitem{10214340}
Z.~Yang, Q.~Xu, W.~Hou, S.~Bao, Y.~He, X.~Cao, and Q.~Huang, ``Revisiting
  auc-oriented adversarial training with loss-agnostic perturbations,''
  \emph{IEEE TPAMI}, 2023.

\bibitem{10478545}
X.~Jia, Y.~Zhang, X.~Wei, B.~Wu, K.~Ma, J.~Wang, and X.~Cao, ``Improving fast
  adversarial training with prior-guided knowledge,'' \emph{IEEE TPAMI}, 2024.

\bibitem{SIA}
X.~Wang, Z.~Zhang, and J.~Zhang, ``Structure invariant transformation for
  better adversarial transferability,'' in \emph{ICCV}, 2023.

\bibitem{BSR}
K.~Wang, X.~He, W.~Wang, and X.~Wang, ``Boosting adversarial transferability by
  block shuffle and rotation,'' in \emph{CVPR}, 2024.

\bibitem{decowa}
Q.~Lin, C.~Luo, Z.~Niu, X.~He, W.~Xie, Y.~Hou, L.~Shen, and S.~Song, ``Boosting
  adversarial transferability across model genus by deformation-constrained
  warping,'' \emph{AAAI}, 2024.

\bibitem{momentum2018dong}
Y.~Dong, F.~Liao, T.~Pang, H.~Su, J.~Zhu, X.~Hu, and J.~Li, ``Boosting
  adversarial attacks with momentum,'' in \emph{CVPR}, 2018.

\bibitem{evading2019dong}
Y.~Dong, T.~Pang, H.~Su, and J.~Zhu, ``Evading defenses to transferable
  adversarial examples by translation-invariant attacks,'' in \emph{CVPR},
  2019.

\bibitem{chen2023selfensemble}
S.~Chen, G.~Yuan, X.~Cheng, Y.~Gong, M.~Qin, Y.~Wang, and X.~Huang,
  ``Self-ensemble protection: Training checkpoints are good data protectors,''
  in \emph{ICLR}, 2023.

\bibitem{ODI}
J.~Byun, S.~Cho, M.-J. Kwon, H.-S. Kim, and C.~Kim, ``Improving the
  transferability of targeted adversarial examples through object-based diverse
  input,'' in \emph{CVPR}, 2022.

\bibitem{9858024}
F.~Yang, J.~Weng, Z.~Zhong, H.~Liu, Z.~Wang, Z.~Luo, D.~Cao, S.~Li, S.~Satoh,
  and N.~Sebe, ``Towards robust person re-identification by defending against
  universal attackers,'' \emph{IEEE Transactions on Pattern Analysis and
  Machine Intelligence}, 2023.

\bibitem{liu2016delving}
Y.~Liu, X.~Chen, C.~Liu, and D.~Song, ``Delving into transferable adversarial
  examples and black-box attacks,'' in \emph{ICLR}, 2017.

\bibitem{PoTrip}
M.~Li, C.~Deng, T.~Li, J.~Yan, X.~Gao, and H.~Huang, ``Towards transferable
  targeted attack,'' in \emph{CVPR}, 2020.

\bibitem{logit}
Z.~Zhao, Z.~Liu, and M.~Larson, ``On success and simplicity: A second look at
  transferable targeted attacks,'' in \emph{NeurIPS}, 2021.

\bibitem{SelfU}
Z.~Wei, J.~Chen, Z.~Wu, and Y.-G. Jiang, ``Enhancing the self-universality for
  transferable targeted attacks,'' in \emph{CVPR}, 2023.

\bibitem{AOSBLL}
X.~Sun, G.~Cheng, H.~Li, L.~Pei, and J.~Han, ``On single-model transferable
  targeted attacks: A closer look at decision-level optimization,'' \emph{IEEE
  TIP}, 2023.

\bibitem{marginangleT}
J.~Weng, Z.~Luo, S.~Li, N.~Sebe, and Z.~Zhong, ``Logit margin matters:
  Improving transferable targeted adversarial attack by logit calibration,''
  \emph{IEEE TIFS}, 2023.

\bibitem{Ilyas19}
A.~Ilyas, S.~Santurkar, D.~Tsipras, L.~Engstrom, B.~Tran, and A.~Madry,
  ``{Adversarial Examples Are Not Bugs, They Are Features},'' in
  \emph{NeurIPS}, 2019.

\bibitem{domainfeatureuap}
C.~Zhang, P.~Benz, T.~Imtiaz, and I.~S. Kweon, ``Understanding adversarial
  examples from the mutual influence of images and perturbations,'' in
  \emph{CVPR}, 2020.

\bibitem{naseer2021generating}
M.~Naseer, S.~Khan, M.~Hayat, F.~S. Khan, and F.~Porikli, ``On generating
  transferable targeted perturbations,'' in \emph{ICCV}, 2021.

\bibitem{ben2006analysis}
S.~Ben-David, J.~Blitzer, K.~Crammer, and F.~Pereira, ``Analysis of
  representations for domain adaptation,'' \emph{NeurIPS}, 2006.

\bibitem{inputdiversity2019xie}
C.~Xie, Z.~Zhang, Y.~Zhou, S.~Bai, J.~Wang, Z.~Ren, and A.~L. Yuille,
  ``Improving transferability of adversarial examples with input diversity,''
  in \emph{CVPR}, 2019.

\bibitem{RDI}
J.~Zou, Z.~Pan, J.~Qiu, X.~Liu, T.~Rui, and W.~Li, ``Improving the
  transferability of adversarial examples with resized-diverse-inputs,
  diversity-ensemble and region fitting,'' in \emph{ECCV}, 2020.

\bibitem{zhao2023minimizing}
A.~Zhao, T.~Chu, Y.~Liu, W.~Li, J.~Li, and L.~Duan, ``Minimizing maximum model
  discrepancy for transferable black-box targeted attacks,'' in \emph{CVPR},
  2023.

\bibitem{weng2023exploring}
J.~Weng, Z.~Luo, Z.~Zhong, D.~Lin, and S.~Li, ``Exploring non-target knowledge
  for improving ensemble universal adversarial attacks,'' in \emph{AAAI}, 2023.

\bibitem{nestrov2019lin}
J.~Lin, C.~Song, K.~He, L.~Wang, and J.~E. Hopcroft, ``Nesterov accelerated
  gradient and scale invariance for adversarial attacks,'' in \emph{ICLR},
  2020.

\bibitem{Yun_2021_CVPR}
S.~Yun, S.~J. Oh, B.~Heo, D.~Han, J.~Choe, and S.~Chun, ``Re-labeling imagenet:
  From single to multi-labels, from global to localized labels,'' in
  \emph{CVPR}, 2021.

\bibitem{admix}
X.~Wang, X.~He, J.~Wang, and K.~He, ``Admix: Enhancing the transferability of
  adversarial attacks,'' in \emph{ICCV}, 2021.

\bibitem{SSA}
Y.~Long, Q.~Zhang, B.~Zeng, L.~Gao, X.~Liu, J.~Zhang, and J.~Song, ``Frequency
  domain model augmentation for adversarial attack,'' in \emph{ECCV}, 2022.

\bibitem{kurakin2016adversarial}
A.~Kurakin, I.~Goodfellow, S.~Bengio \emph{et~al.}, ``Adversarial examples in
  the physical world,'' in \emph{ICLR}, 2017.

\bibitem{CFM}
J.~Byun, M.-J. Kwon, S.~Cho, Y.~Kim, and C.~Kim, ``Introducing competition to
  boost the transferability of targeted adversarial examples through clean
  feature mixup,'' in \emph{CVPR}, 2023.

\bibitem{inkawhich2020perturbing}
N.~Inkawhich, K.~Liang, B.~Wang, M.~Inkawhich, L.~Carin, and Y.~Chen,
  ``Perturbing across the feature hierarchy to improve standard and strict
  blackbox attack transferability,'' in \emph{NeurIPS}, 2020.

\bibitem{wang2023towards}
Z.~Wang, H.~Yang, Y.~Feng, P.~Sun, H.~Guo, Z.~Zhang, and K.~Ren, ``Towards
  transferable targeted adversarial examples,'' in \emph{CVPR}, 2023.

\bibitem{naseer2022stylized}
M.~Naseer, S.~Khan, M.~Hayat, F.~S. Khan, and F.~Porikli, ``Stylized
  adversarial defense,'' \emph{IEEE TPAMI}, 2022.

\bibitem{yuan2022adaptive}
Z.~Yuan, J.~Zhang, and S.~Shan, ``Adaptive image transformations for
  transfer-based adversarial attack,'' in \emph{ECCV}, 2022.

\bibitem{Zhu_2024_CVPR}
R.~Zhu, Z.~Zhang, S.~Liang, Z.~Liu, and C.~Xu, ``Learning to transform
  dynamically for better adversarial transferability,'' in \emph{CVPR}, 2024.

\bibitem{resnet2016kh}
K.~He, X.~Zhang, S.~Ren, and J.~Sun, ``Deep residual learning for image
  recognition,'' in \emph{CVPR}, 2016.

\bibitem{densenet2017huang}
G.~Huang, Z.~Liu, L.~van~der Maaten, and K.~Q. Weinberger, ``Densely connected
  convolutional networks,'' in \emph{CVPR}, 2017.

\bibitem{russakovsky2015imagenet}
O.~Russakovsky, J.~Deng, H.~Su, J.~Krause, S.~Satheesh, S.~Ma, Z.~Huang,
  A.~Karpathy, A.~Khosla, M.~Bernstein \emph{et~al.}, ``Imagenet large scale
  visual recognition challenge,'' \emph{IJCV}, 2015.

\bibitem{Sandler_2018_CVPR}
M.~Sandler, A.~Howard, M.~Zhu, A.~Zhmoginov, and L.-C. Chen, ``Mobilenetv2:
  Inverted residuals and linear bottlenecks,'' in \emph{CVPR}, 2018.

\bibitem{efficientnet}
M.~Tan and Q.~Le, ``{E}fficient{N}et: Rethinking model scaling for
  convolutional neural networks,'' in \emph{ICML}, 2019.

\bibitem{liu2022convnet}
Z.~Liu, H.~Mao, C.-Y. Wu, C.~Feichtenhofer, T.~Darrell, and S.~Xie, ``A convnet
  for the 2020s,'' in \emph{CVPR}, 2022.

\bibitem{szegedy2016rethinking}
C.~Szegedy, V.~Vanhoucke, S.~Ioffe, J.~Shlens, and Z.~Wojna, ``Rethinking the
  inception architecture for computer vision,'' in \emph{CVPR}, 2016.

\bibitem{szegedy2017inception}
C.~Szegedy, S.~Ioffe, V.~Vanhoucke, and A.~Alemi, ``Inception-v4,
  inception-resnet and the impact of residual connections on learning,'' in
  \emph{AAAI}, 2017.

\bibitem{chollet2017xception}
F.~Chollet, ``Xception: Deep learning with depthwise separable convolutions,''
  in \emph{CVPR}, 2017.

\bibitem{dosovitskiy2020image}
A.~Dosovitskiy, L.~Beyer, A.~Kolesnikov, D.~Weissenborn, X.~Zhai,
  T.~Unterthiner, M.~Dehghani, M.~Minderer, G.~Heigold, S.~Gelly, J.~Uszkoreit,
  and N.~Houlsby, ``An image is worth 16x16 words: Transformers for image
  recognition at scale,'' in \emph{ICLR}, 2021.

\bibitem{Liu_2021_ICCV}
Z.~Liu, Y.~Lin, Y.~Cao, H.~Hu, Y.~Wei, Z.~Zhang, S.~Lin, and B.~Guo, ``Swin
  transformer: Hierarchical vision transformer using shifted windows,'' in
  \emph{ICCV}, 2021.

\bibitem{tu2022maxvit}
Z.~Tu, H.~Talebi, H.~Zhang, F.~Yang, P.~Milanfar, A.~Bovik, and Y.~Li,
  ``Maxvit: Multi-axis vision transformer,'' in \emph{ECCV}, 2022.

\bibitem{chu2021twins}
X.~Chu, Z.~Tian, Y.~Wang, B.~Zhang, H.~Ren, X.~Wei, H.~Xia, and C.~Shen,
  ``Twins: Revisiting the design of spatial attention in vision transformers,''
  in \emph{NeurIPS}, 2021.

\bibitem{heo2021rethinking}
B.~Heo, S.~Yun, D.~Han, S.~Chun, J.~Choe, and S.~J. Oh, ``Rethinking spatial
  dimensions of vision transformers,'' in \emph{ICCV}, 2021.

\bibitem{NEURIPS2021_854d9fca}
K.~Han, A.~Xiao, E.~Wu, J.~Guo, C.~XU, and Y.~Wang, ``Transformer in
  transformer,'' in \emph{NeurIPS}, 2021.

\bibitem{pmlr-v139-touvron21a}
H.~Touvron, M.~Cord, M.~Douze, F.~Massa, A.~Sablayrolles, and H.~Jegou,
  ``Training data-efficient image transformers \& distillation through
  attention,'' in \emph{ICML}, 2021.

\bibitem{rw2019timm}
R.~Wightman, ``Pytorch image models,''
  \url{https://github.com/rwightman/pytorch-image-models}, 2019.

\bibitem{NEURIPS2019_bdbca288}
A.~Paszke, S.~Gross, F.~Massa, A.~Lerer, J.~Bradbury, G.~Chanan, T.~Killeen,
  Z.~Lin, N.~Gimelshein, L.~Antiga, A.~Desmaison, A.~Kopf, E.~Yang, Z.~DeVito,
  M.~Raison, A.~Tejani, S.~Chilamkurthy, B.~Steiner, L.~Fang, J.~Bai, and
  S.~Chintala, ``Pytorch: An imperative style, high-performance deep learning
  library,'' in \emph{NeurIPS}, 2019.

\bibitem{hendrycks2019augmix}
D.~Hendrycks, N.~Mu, E.~D. Cubuk, B.~Zoph, J.~Gilmer, and B.~Lakshminarayanan,
  ``Augmix: A simple method to improve robustness and uncertainty under data
  shift,'' in \emph{ICLR}, 2020.

\bibitem{geirhos2018imagenet}
R.~Geirhos, P.~Rubisch, C.~Michaelis, M.~Bethge, F.~A. Wichmann, and
  W.~Brendel, ``Imagenet-trained {CNN}s are biased towards texture; increasing
  shape bias improves accuracy and robustness.'' in \emph{ICLR}, 2019.

\bibitem{vgg2015sk}
K.~Simonyan and A.~Zisserman, ``Very deep convolutional networks for
  large-scale image recognition,'' in \emph{ICLR}, 2015.

\bibitem{salman2020adversarially}
H.~Salman, A.~Ilyas, L.~Engstrom, A.~Kapoor, and A.~Madry, ``Do adversarially
  robust imagenet models transfer better?'' in \emph{NeurIPS}, 2020.

\end{thebibliography}
	
	\vfill
\end{document}